\documentclass[11pt,a4paper,most]{article}
\usepackage{times,latexsym}
\usepackage{url}
\usepackage[T1]{fontenc}

\usepackage{booktabs}
\usepackage{graphicx}
\usepackage{amsmath}
\usepackage{amssymb}
\usepackage{tabularx}
\usepackage{tcolorbox}
\tcbuselibrary{skins, breakable}
\usepackage{float}
\usepackage{placeins}
\usepackage{makecell}
\usepackage{multirow}
\usepackage{pgfplots}
\usepackage{afterpage}
\usepackage[font=small]{caption}
\usepackage{subcaption}
\usepackage{caption}
\usepackage{float}
\usepackage[acceptedWithA]{tacl2021v1}

\usepackage{xspace,mfirstuc,tabulary}

\newif\iftaclinstructions
\taclinstructionsfalse 
\iftaclinstructions
\renewcommand{\confidential}{}
\renewcommand{\anonsubtext}{(No author info supplied here, for consistency with
TACL-submission anonymization requirements)}
\newcommand{\instr}
\fi

\iftaclpubformat 

\else

\fi

\title{Think Like a Doctor: Conversational Diagnosis \\ through the Exploration of Diagnostic Knowledge Graphs}

\author{Jeongmoon Won$^*$ ~~ Seungwon Kook$^*$ ~~ Yohan Jo$^{\dag}$ \\
  Graduate School of Data Science, Seoul National University \\
  \texttt{\{jmwon06231,cacu5615,yohan.jo\}@snu.ac.kr} \\
}

\begin{document}
\maketitle
\def\thefootnote{\fnsymbol{footnote}}
\footnotetext[1]{Equal contribution.}
\footnotetext[2]{Corresponding author.}
\def\thefootnote{\arabic{footnote}}

\begin{abstract}
Conversational diagnosis requires multi-turn history-taking, where an agent asks clarifying questions to refine differential diagnoses under incomplete information. Existing approaches often rely on the parametric knowledge of a model or assume that patients provide rich and concrete information, which is unrealistic. To address these limitations, we propose a conversational diagnosis system that explores a diagnostic knowledge graph to reason in two steps: (i) generating diagnostic hypotheses from the dialogue context, and (ii) verifying hypotheses through clarifying questions, which are repeated until a final diagnosis is reached. Since evaluating the system requires a realistic patient simulator that responds to the system's questions, we adopt PatientSim, a persona-driven patient simulator, together with patient profiles from MIMIC-IV. We further adapt it with low-specificity symptom reporting to reflect how real-world patients describe symptoms vaguely during early clinical encounters. Experiments show improved diagnostic accuracy and efficiency over strong baselines, and physician evaluations support the realism of our simulator and the clinical utility of the generated clarifying questions. Our code will be released upon publication.
\end{abstract}

\section{Introduction}

Clinical diagnosis is inherently a multi-turn process of history-taking. In real clinical encounters, clinicians interact with patients to elicit missing information, refine differential diagnoses, and decide which question to ask next under uncertainty. Although Large Language Models (LLMs) have shown strong performance on medical benchmarks \citep{ijcai2025p1169, nori2025sequential, singhal2025toward}, such benchmarks typically evaluate medical knowledge in settings where relevant information is provided upfront. They therefore do not fully capture the interactive nature of clinical diagnosis, where the central challenge is not merely interpreting given facts but proactively and efficiently gathering diagnostically relevant evidence across turns.

Conversational diagnosis systems can serve as decision-support tools for outpatient initial consultations, where history-taking is often the primary source of diagnostic evidence. A major obstacle, however, is that patients often describe symptoms in vague, incomplete, or underspecified terms (e.g., ``my belly feels weird'') \citep{redelmeier2001problems, dahm2023communication, gaber2025evaluating}, which increases the risk of misdiagnosis. Diagnostic accuracy therefore depends on asking the right \emph{clarifying questions (CQs)}—those that effectively target uncertainty and elicit diagnostically relevant cues. As illustrated in Figure~\ref{fig:intro_concept}, directly answering from an initial underspecified utterance often leads to generic responses (Figure~\ref{fig:intro_a}), whereas iterative clarification through CQs enables more precise diagnosis (Figure~\ref{fig:intro_b}).

\afterpage{
\clearpage
\begin{figure}[t]
    \centering
    \begin{subfigure}[b]{\columnwidth}
        \centering
        \includegraphics[width=0.9\textwidth]{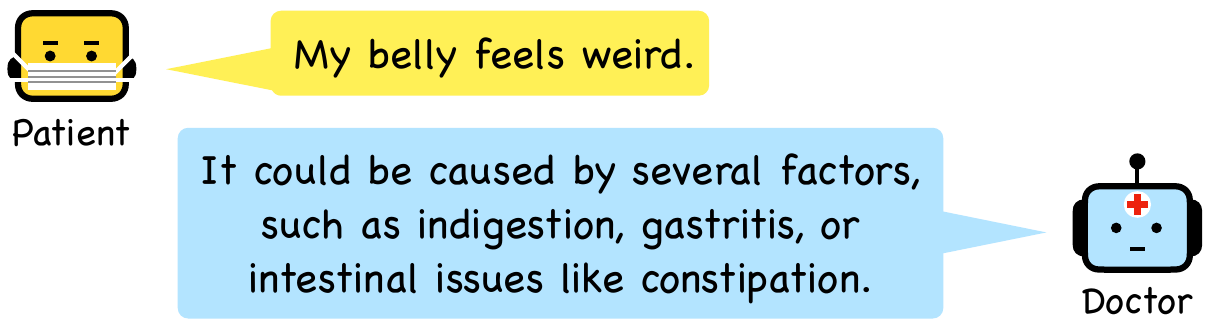}
        \caption{Direct answer with generic information.}
        \label{fig:intro_a}
    \end{subfigure}
    
    \vspace{0.5em}
    
    \begin{subfigure}[b]{\columnwidth}
        \centering
        \includegraphics[width=0.9\textwidth]{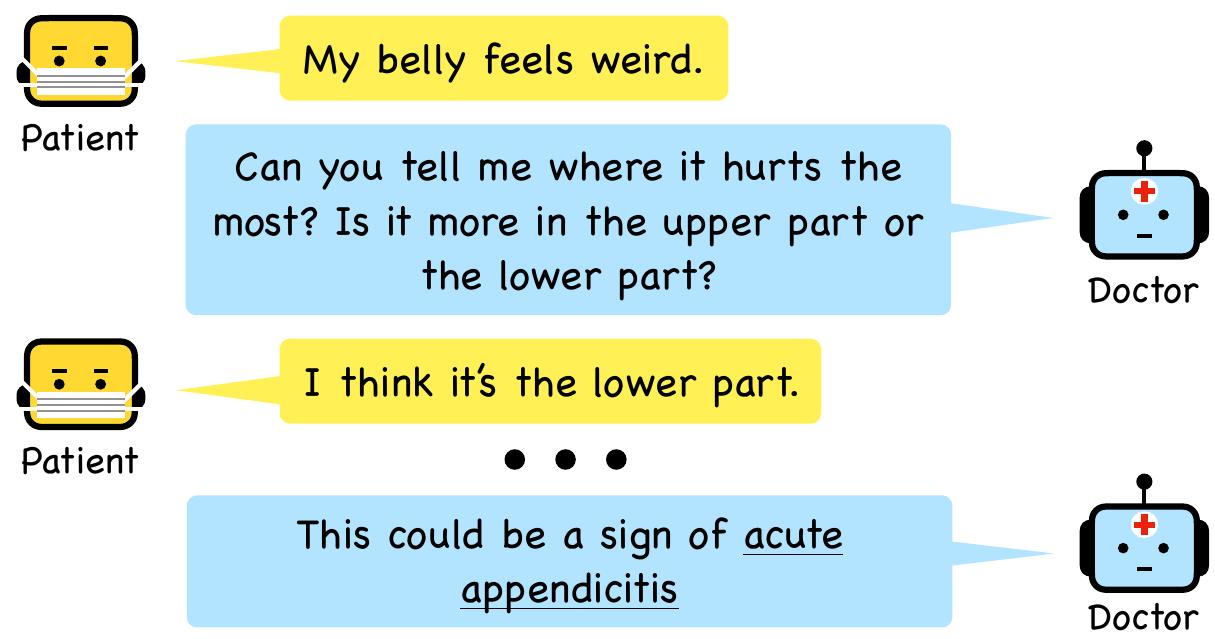}
        \caption{History-taking through clarifying questions.}
        \label{fig:intro_b}
    \end{subfigure}
    
    \caption{Conceptual comparison between two diagnostic settings.}
    \label{fig:intro_concept}
\end{figure}
}

Recent work has explored multi-turn conversational diagnosis systems with CQs, moving beyond static medical QA toward interactive history-taking. However, important limitations remain. Existing systems often lack explicit grounding in structured diagnostic knowledge, making it unclear whether their CQs are diagnostically justified. They may also assume patient responses that are richer, more concrete, or more constrained than those observed in early-stage clinical encounters. These assumptions are misaligned with realistic history-taking, where patients reveal information gradually and often begin with vague or incomplete symptom descriptions.

Clinical reasoning literature describes diagnosis as a hypothesis-driven process in which clinicians form plausible diagnostic hypotheses from early cues and gather additional information to discriminate among them \citep{yazdani2017models, croskerry2009universal, norman2024dual}. Motivated by this view, we operationalize the information-gathering step as targeted CQs that verify or rule out candidate diseases. This process requires diagnostic knowledge that organizes which cues are informative for distinguishing among competing hypotheses. To support such reasoning in a conversational system, we represent diagnostic knowledge as a \textit{diagnostic knowledge graph} that enables explicit cross-disease comparison.



Accordingly, we propose a conversational diagnosis system that operates in two steps: (i) generating plausible hypotheses and extracting a hypothesis-driven subgraph from the diagnostic knowledge graph, and (ii) using the accumulated evidence and dialogue history to generate either a CQ or a final diagnosis.
To assess whether the system can actively compensate for underspecified patient utterances through targeted CQs, we build on PatientSim, a persona-driven simulator for realistic doctor-patient interactions, and augment it with low-specificity symptom reporting. This allows us to evaluate the system under patient behavior that is both persona-varied and symptom-underspecified. 

Empirically, we find that grounding conversational diagnosis in the diagnostic knowledge graph leads to more accurate and efficient diagnosis. Specifically, grounding diagnosis in the graph substantially improves performance over relying on parametric knowledge alone. Moreover, generating CQs based on a small set of plausible hypotheses leads to more targeted and informative CQs. Models trained under this system also achieve stronger diagnostic performance with fewer dialogue turns, indicating more focused and efficient history-taking. Furthermore, two physician evaluations support these findings---one assessing our patient simulator's realism and one assessing the clinical utility of the generated clarifying questions---while also surfacing concrete limitations, which we report with nuance. These findings suggest that hypothesis-driven graph grounding provides a practical design principle for building reliable conversational diagnosis systems. Our system can potentially support preliminary diagnosis and patient triage in time-constrained clinical settings, as well as assist clinicians in asking critical questions under heavy workload.

\begin{figure*}[t]
    \centering
    \includegraphics[width=\textwidth]{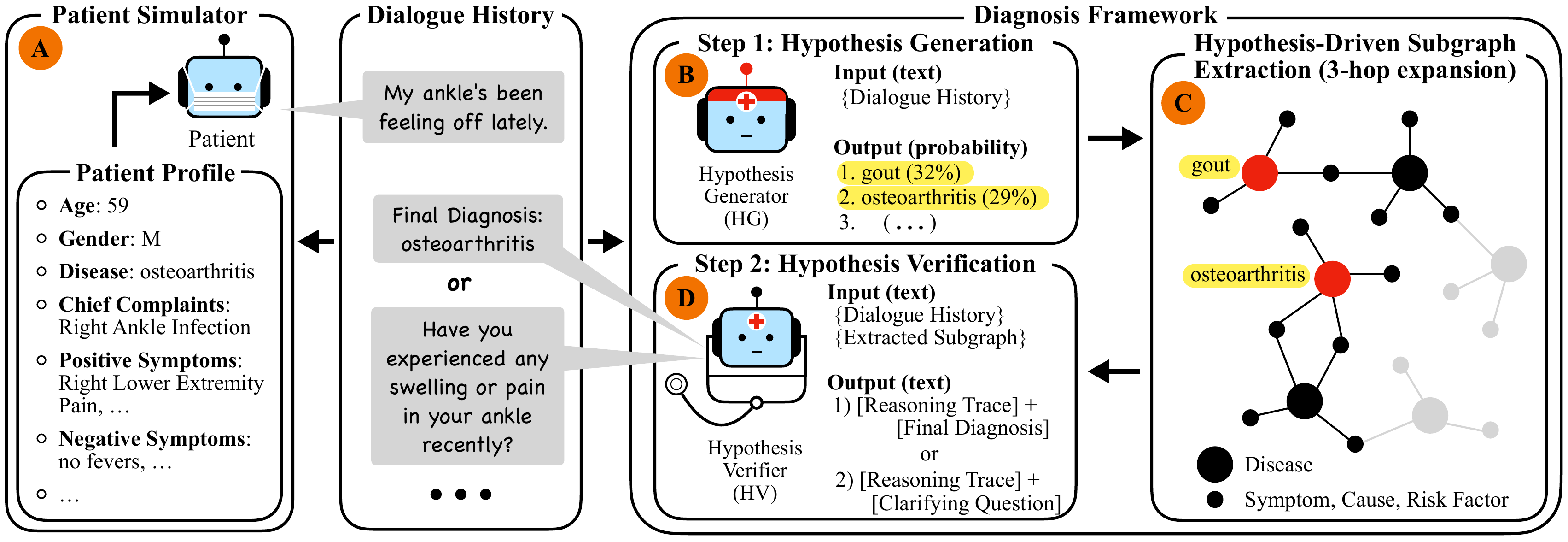}
    \caption{Our system (i) generates diagnostic hypotheses from the dialogue context (\S{\ref{subsubsec:HG}}), and (ii) verifies hypotheses through clarifying questions, which are repeated until the final diagnosis (\S{\ref{subsubsec:HV}}).
    }
    \label{fig:framework_overview}
\end{figure*}

\section{Related Work}
\subsection{Generating Clarifying Questions in Conversational Diagnosis}
Prior work has explored conversational diagnosis as a multi-turn process in which a system asks CQs to collect missing evidence before diagnosing. However, we found two recurring limitations. The first limitation concerns assumptions about patient behavior. Some retrieval-augmented approaches implicitly assume that patients provide rich and informative context upfront, thereby overlooking the reality of early-stage history-taking, where patient descriptions are often vague or underspecified \citep{Sun2025MRDRAG, Xu2024IADDx}. Other policy-optimization approaches simplify interaction by restricting patient responses to three choices (e.g., yes/no/unknown) \citep{jia2025ddo, sun2024conversational}. This setting requires an excessive number of granular CQs that must be answerable with yes/no (e.g., asking for a location via separate queries for upper, lower, left, and right). In contrast, our system is designed to operate under vague or underspecified patient utterances.

The second limitation concerns how systems leverage medical knowledge. Some recent fine-tuned or prompting-based models primarily rely on parametric knowledge without explicit external grounding, making their CQs and diagnoses difficult to verify \citep{lai2025doctorr1, feng2025doctoragentrl, MEDIQ}. Other work incorporates knowledge graphs, but often uses them only for disease scoring or entity-level retrieval rather than for guiding discriminative questioning \citep{chen2025cod, MedKP}. Our system addresses this gap by explicitly grounding CQ generation and diagnosis decisions in structured diagnostic knowledge.

\subsection{Patient Simulator}
Conversational diagnosis systems commonly rely on patient simulators to enable scalable multi-turn training and evaluation. Prior simulators have improved realism by grounding patient responses in structured patient profiles \citep{liu2024medchain}, and by modeling demographic attributes, personality traits, dialogue states, etc \citep{liao2024automatic, bao2025sfmss, liu2025exploring}. PatientSim  \citep{kyung2025patientsim} extends this line of work by simulating diverse patient personas encountered in clinical settings, including variations in language proficiency, personality, recall, and confusion. However, these persona dimensions mainly control how patients communicate or remember information, whereas early-stage diagnostic history-taking also depends on how specifically patients describe their current symptoms. Real patients frequently provide incomplete, noisy, or low-specificity descriptions and may omit diagnostically relevant details unless explicitly prompted \citep{redelmeier2001problems, dahm2023communication, gaber2025evaluating}. We therefore build on PatientSim and introduce low-specificity symptom description as a new trait, enabling more realistic evaluation of whether CQs can elicit missing diagnostic details from vague patient utterances.

\section{Methodology}
\label{sec:methodology}
\subsection{Problem Definition}
Let $\mathcal{D} = [(u_1^P, u_1^D), \dots, (u_T^P, u_T^D)]$ denote a dialogue session, where $u_t^P$ and $u_t^D$ represent the patient utterance and system response at turn $t$, respectively. The system response $u_t^D$ takes the form of either a CQ or a final diagnosis. We define the dialogue history up to turn $t$ as $H_t = (u_1^P, u_1^D, \dots, u_t^P)$. Given $H_t$ and a diagnostic knowledge graph $\mathcal{G}$, our objective is to generate an optimal response $u_t^D$ that minimizes diagnostic uncertainty and leads to the correct identification of the ground-truth disease $d_{gold}$. Figure~\ref{fig:framework_overview} provides an overview of our system.

\subsection{Diagnostic Knowledge Graph}
We construct a diagnostic knowledge graph $\mathcal{G}$ that provides structured medical grounding to support diagnostic hypothesis generation and verification through CQs. While existing biomedical knowledge graphs \citep{bodenreider2004umls, chandak2023building} offer broad and expert-level medical knowledge, their wide scope often introduces extraneous information that acts as noise during diagnostic reasoning \citep{zhai-etal-2024-towards, mavi-saparov-zhao-2023-retrieval}. Instead, we derive $\mathcal{G}$ from diagnostic schemas—clinical decision trees commonly used by clinicians and described in standard medical textbooks. However, directly using such schemas is ill-suited for conversational diagnosis, because they organize information separately for each disease. As a result, a system cannot easily reason from a single symptom to the set of competing diseases associated with it, nor compare those diseases to determine which CQs are most discriminative.

To address this limitation, we reformulate disease-specific schemas into a unified graph representation that explicitly supports cross-disease comparison. In this graph, multiple diseases share attribute nodes (\texttt{symptom}, \texttt{cause}, and \texttt{risk\_factor}), capturing many-to-many relationships between diseases and clinical cues. This shared structure enables the system to reason over overlapping and contrasting attributes among candidate diseases, directly supporting discriminative CQ generation during history-taking. The resulting $\mathcal{G}$ contains 338 disease nodes connected to attribute nodes via 3,935 edges. Detailed information about our diagnostic schemas and graph is provided in Appendix~\ref{app:kg_details}.

\subsection{Diagnosis Framework}
We model conversational diagnosis as an iterative, hypothesis-driven reasoning process in which the system alternates between generating diagnostic hypotheses and verifying them through CQs.

\subsubsection{Step 1: Hypothesis Generation (Figure~\ref{fig:framework_overview}, B)}
\label{subsubsec:HG}
The goal of this step is to generate a small set of plausible diagnostic hypotheses from the current dialogue history. Each hypothesis corresponds to a candidate disease node in $\mathcal{G}$ and serves as an anchor for subsequent evidence-driven questioning. By first identifying plausible diseases, the system can focus its CQs on the most useful information for hypothesis verification.

Given the dialogue history $H_t$, we predict probability scores over all diseases defined in $\mathcal{G}$. We formulate this prediction as a multi-label classification task using a \emph{Hypothesis Generator (HG)}. HG takes $H_t$ as input and computes:
\begin{equation}
\resizebox{0.88\columnwidth}{!}{$
    P(\text{Disease} \mid H_t) = \sigma \bigl( W \cdot \mathrm{LLM}_{\text{HG}}(H_t) + b \bigr),
$}
\end{equation}
where $\mathrm{LLM}_{\text{HG}}$ denotes the underlying language model, $W$ and $b$ are the learnable weight matrix and bias of the classification head applied to the final hidden state of the last token, and $\sigma$ is the sigmoid function. Based on these scores, we select the top-$n$ diseases with the highest probabilities. The prompt template used for HG is in Appendix~\ref{app:classifer_prompts}

\paragraph{Hypothesis-Driven Subgraph Extraction. (Figure~\ref{fig:framework_overview}, C)}
The selected top-$n$ diseases are treated as anchor hypotheses (represented by red nodes in Figure~\ref{fig:framework_overview}, C). Relying on anchor hypotheses alone is insufficient for making a diagnosis. At this stage, the predicted probabilities have not yet been verified against concrete clinical evidence: the patient may not exhibit the detailed attributes associated with the anchor hypotheses, and the true diagnosis may be a different disease that shares the same attribute nodes. Effective hypothesis verification therefore requires structured access to the attributes that connect competing diseases and differentiate them from plausible alternatives. To this end, we extract a localized subgraph $\mathcal{G}_{\text{sub}}$ that provides the necessary diagnostic context.

We perform a 3-hop expansion starting from the anchor disease nodes. Specifically, the expansion includes: (1) 1-hop: to their direct attributes (\texttt{symptom}, \texttt{cause}, \texttt{risk\_factor}); (2) 2-hop: to other diseases sharing these attributes, representing competing hypotheses (large black nodes in Figure~\ref{fig:framework_overview}, C); and (3) 3-hop: to the attributes of those competing diseases. By exposing both shared and disease-specific attributes, this structure enables the system to generate CQs that either strengthen confidence in a plausible hypothesis or rule it out by probing discriminative attributes. To control the size of $\mathcal{G}_{\text{sub}}$, we filter the competing diseases by retaining only those whose predicted probabilities exceed a threshold $\tau$.

\subsubsection{Step 2: Hypothesis Verification (Figure~\ref{fig:framework_overview}, D)}
\label{subsubsec:HV}
In the second step, the \emph{Hypothesis Verifier (HV)} validates the candidate diseases identified in Step~1 by actively collecting missing evidence. Given the dialogue history $H_t$ and the extracted subgraph $\mathcal{G}_{\text{sub}}$, the system determines whether to ask a CQ or provide a final diagnosis.

We linearize $\mathcal{G}_{\text{sub}}$ into explicit textual statements (e.g., ``Disease A causes Symptom B'') and incorporate them into the HV's prompt context. This design enables explicit and interpretable reasoning over diagnostic knowledge without the need to train additional graph encoders, thereby improving generality and ease of integration.

The HV then generates an explicit reasoning trace $r_t$ and a response $u_t^D$:
\begin{equation}
r_t,\; u_t^D = \mathrm{LLM}_{\text{HV}}\!\left(H_t,\, \text{Text}(\mathcal{G}_{\text{sub}})\right).
\end{equation}
If the HV judges the accumulated evidence insufficient, it generates a CQ grounded in $\mathcal{G}_{\text{sub}}$ to confirm or rule out specific attributes. Conversely, if the evidence sufficiently supports a hypothesis, the HV outputs a final diagnosis. This process iterates, with the patient simulator providing the next utterance $u_{t+1}^P$, until a diagnosis is reached. The prompt template used for HV is in Appendix~\ref{app:doctor_prompts}.

\subsection{Patient Simulator (Figure~\ref{fig:framework_overview}, A)}
\label{sec:patient_simulator}
In our diagnosis framework, we introduce a realistic patient simulator to play the role of a patient that responds to the system’s CQs. 

At each turn, 
the simulator takes the underlying patient profile and $u_t^D$ as input, and generates a response $u_t^P$. We keep the input template unchanged to ensure a controlled comparison with the baseline PatientSim, isolating the effect of low-specificity symptom reporting.
When the queried information is available in the profile, the simulator provides an answer; otherwise, it indicates that the information is uncertain or unknown. 
Because the simulator answers from a pre-structured profile with explicit fields rather than from a raw clinical record, answering a query is a bounded lookup over labeled fields rather than open-ended retrieval. To quantify the risk of fabricated answers, three annotators independently audited the full set of 45 evaluation dialogues (265 simulator turns) against the labeled profile fields. Only $0.75\%$ of turns contradicted the underlying profile, with high inter-annotator agreement (Gwet's AC1 $=0.98$).


\subsubsection{Data Source and Preprocessing}
\label{subsubsec:data_preprocessing}
Patient profiles are constructed from the MIMIC-IV dataset suite \citep{johnson2024mimiciv, johnson2024mimiciv_ed, johnson2023mimiciv_note}, a large-scale, de-identified electronic health record dataset. We primarily utilize MIMIC-IV-ED (Emergency Department) \citep{johnson2024mimiciv_ed}, which contains visit-level information such as chief complaints and encounter diagnoses. We augment it with MIMIC-IV-Note \citep{johnson2023mimiciv_note} to aggregate clinical notes required for constructing detailed patient profiles. Emergency department encounters closely resemble time-constrained initial consultations under incomplete information. This makes our system well-suited to support efficient history-taking and indispensable for reducing patient wait times through automated preliminary diagnosis and patient triage. 

Each profile includes (i) basic information (age, gender, chief complaint, and disease labels), (ii) history of present illness, including symptoms asked about during the encounter and whether the patient affirmed or denied them, and (iii) background information such as past medical history, family history, and social history. While most cues (e.g., current symptoms) are readily reportable by patients, some details such as precise medical history may be uncertain; this behavior is modeled through persona and specificity traits as described in the following section (see \S{\ref{subsubsec:simulator_augmentation}}).

For these patient profiles, we first determine which MIMIC labels correspond to diagnosable diseases. Across the 50{,}289 MIMIC-IV-ED profiles, 4{,}937 unique diagnosis labels appear, comprising 2{,}244 disease and 2{,}693 non-disease labels---the latter being clinical findings rather than well-defined disease entities (e.g., abdominal pain, syncope). This leaves 20{,}672 profiles with a genuine disease label. We then align these disease labels with disease nodes in $\mathcal{G}$: we first apply exact string matching, and when an exact match is unavailable, we perform LLM-based semantic matching to map clinically equivalent disease names. Each profile contains between one and four diagnosed diseases. Because our system grounds reasoning in $\mathcal{G}$, we retain only profiles whose disease labels can be mapped to graph nodes. 

Our diagnostic knowledge graph covers 15{,}252 of these profiles (${\sim}74\%$ of the diagnosable cases), while the remaining 5{,}420 (${\sim}26\%$) involve diseases outside the original graph. The retained profiles still cover major disease categories, including respiratory (e.g., upper respiratory tract infection), gastrointestinal (e.g., acute appendicitis), and dermatologic conditions (e.g., cellulitis). The retained and filtered cohorts are demographically near-identical (age, gender, race, insurance, and language), so filtering reflects the graph's clinical-category scope rather than patient-selection bias. We therefore further examine graph expansion by adding these out-of-graph diseases and evaluating the resulting system in \S\ref{sec:ood} and Appendix~\ref{app:graph_expansion}.


To mitigate class imbalance, we perform stratified sampling based on disease types and co-occurrence counts when creating train, validation, and test splits. We further limit the number of training profiles to 60 per disease. Dataset statistics are reported in Table~\ref{tab:patient_stats}.

\begin{table}[t]
    \centering
    \small
    \begin{tabular}{lccc}
        \toprule
        \textbf{Component} & \textbf{Train} & \textbf{Valid} & \textbf{Test} \\
        \midrule
        Hypothesis Generator & 1,390 & 172 & 174 \\
        Hypothesis Verifier & 1,449 & 304 & 275 \\
        \bottomrule
    \end{tabular}
    \caption{The number of patient profiles used for training and validation of the HG and HV. The test split is used exclusively for model selection among alternative HG variants.}
    \label{tab:patient_stats}
\end{table}

\subsubsection{Simulator Adoption and Specificity Augmentation}
\label{subsubsec:simulator_augmentation}
We adopt PatientSim \citep{kyung2025patientsim}, the persona-driven framework for simulating a patient, as the backbone of our simulator. PatientSim modulates four dimensions of patient behavior: language proficiency, personality, recall level, and confusion level. Each simulated patient is instantiated as a combination of attributes across these dimensions. For example, low proficiency yields short, fragmented utterances; distrustful personas hesitate to share information; low recall leads to uncertain medical history; high confusion produces tangential or inconsistent responses.

In PatientSim, patient responses often contain sufficient diagnostic details. In real clinical encounters, however, patients frequently fail to provide enough detail, and their symptom descriptions are underspecified \citep{redelmeier2001problems, dahm2023communication}. To better reflect this behavior, we instruct the simulator to describe symptoms with low specificity along five symptom-description attributes: \textit{location}, \textit{character}, \textit{duration}, \textit{onset}, and \textit{aggravating/relieving factors}.

To ensure that low-specificity behavior combines naturally with existing personas, we introduce additional prompt components informed by interviews with physicians. Specifically, we (i) restrict the \textit{high recall} persona to past medical history while keeping current symptom descriptions less specific, (ii) allow patients with high language proficiency to exhibit a tendency toward self-diagnosis rather than providing detailed symptom descriptions, and (iii) allow \textit{verbose} personas to mask missing symptom details with long but largely irrelevant text. Prompt templates are provided in Appendix~\ref{app:patient_prompts}. In \S{\ref{sec:patient_simulator_eval}}, we empirically verify that this augmented simulator produces patient behaviors that clinicians perceive as more realistic than the original PatientSim.

\subsection{Model Training}
\subsubsection{Synthetic Dialogue Data Generation} 
To train the HG and the HV, we need dialogues that reflect iterative history-taking. Therefore, we construct a synthetic dialogues conditioned on each patient profile in the training and validation set. We generate these dialogues by using GPT-4o-mini \citep{openai2024gpt4omini} via Azure OpenAI Service, in compliance with PhysioNet’s credentialed data use agreement.\footnote{\url{https://physionet.org/news/post/gpt-responsible-use}} We employ two separate instances of GPT-4o-mini to simulate the interaction: one plays the role of the clinician to generate reasoning traces and actions (CQs or diagnoses), while the other serves as the patient simulator (\S\ref{sec:patient_simulator}) when generating patient responses.

During generation, the clinician model is conditioned on the dialogue history and an oracle subgraph $\mathcal{G}_{\text{sub\_oracle}}$, which is constructed by performing a 3-hop expansion from $d_{gold}$. We also provide $d_{gold}$ to ensure medical plausibility during dialogue generation. Importantly, the model is instructed to behave as if the diagnosis is initially unknown and to narrow down hypotheses through iterative questioning.

Although grounding reasoning in $\mathcal{G}$ is central to our system, we do not assume that the model should rely exclusively on the graph. In practice, diagnostic reasoning often requires combining external knowledge with the model’s parametric medical knowledge, especially when patient-provided information is incomplete or weakly covered by the graph. To encourage this adaptive behavior during training, we vary how strongly the model is instructed to rely on the oracle subgraph versus its internal medical knowledge, based on how much of the patient profile is covered by the graph. This allows the model to learn when graph grounding is informative and when flexible use of internal knowledge is necessary. Implementation details are provided in Appendix~\ref{app:synthetic_dialog_gen}.

\subsubsection{Training Details}
\label{sec:training_details}
\paragraph{Hypothesis Generator (HG).}
The HG estimates plausible disease candidates from the current dialogue context. We formulate this as a multi-label classification task, since each patient profile contains between one and four diseases, and use the profile disease labels as ground-truth targets. We initialize $\mathrm{LLM}_{\text{HG}}$ with Qwen-2.5-7B-Instruct \citep{qwen2025qwen25technicalreport} and attach a linear classification head to the final hidden state of the last token, mapping to 338 disease labels. At inference time, we rank diseases by predicted probabilities and select the top-$n$ candidates. 

We also tested alternative hypothesis generation strategies, including (i) a retrieval-based approach that ranks diseases by embedding similarity to patient-mentioned symptoms and (ii) a generative approach that outputs candidate disease names from the dialogue history. We found the classification-head HG to be more effective. Detailed comparisons and experimental settings are provided in Appendix~\ref{app:HG_comparison}. 

To support HG at different dialogue stages, we train it on both full dialogues and their truncated ones.  Specifically, for each full dialogue, we create multiple sub-dialogues by randomly truncating at intermediate turns. The number of variants is set to 20\% of the total dialogue length, and truncation points are sampled uniformly at random. This strategy exposes HG to partial histories of varying lengths, allowing it to learn hypothesis generation under incomplete evidence.

\paragraph{Hypothesis Verifier (HV).}
The HV generates reasoning trace and response at each turn. $\mathrm{LLM}_{\text{HV}}$ is initialized with Qwen-2.5-7B-Instruct and trained via supervised fine-tuning (SFT) on the generated synthetic dialogues. Unlike the HG, the HV is trained on full dialogues without truncation. 

The input includes the instruction, the dialogue history $H_t$, and the extracted subgraph $\mathcal{G}_{\text{sub}}$ for the current turn. The output is structured with special tokens: \texttt{<think>} [Reasoning Trace] \texttt{</think>} followed by either \texttt{<question>} [CQ] \texttt{</question>} or \texttt{<diagnosis>} [Final Diagnosis] \texttt{</diagnosis>}. This format enforces explicit reasoning before committing to a question or diagnosis. Example outputs are shown in Appendix~\ref{app:doctor_response_examples}.

\section{Experiment Settings}
\subsection{Baselines}\label{subsec:baselines} To comprehensively evaluate the effectiveness of our proposed system, we compare it against general-purpose foundation models. In addition, we implemented an adapted version of the Chain-of-Diagnosis (CoD) pipeline \citep{chen2025cod} to benchmark our method against an existing retrieval-based diagnostic framework. 

\paragraph{General-Purpose Models.} We evaluate performance across general-purpose models: GPT-4.1-mini \citep{openai2025gpt4.1mini}, Claude-3.5-Sonnet \citep{anthropic2024claude35sonnet}, Gemini-2.0-Flash \citep{google2024gemini20}, Llama-3.3-70B \citep{grattafiori2024llama3herdmodels}, Qwen-2.5-72B-Instruct  \citep{qwen2025qwen25technicalreport} and Qwen-2.5-7B-Instruct \citep{qwen2025qwen25technicalreport}.

\paragraph{Adapted CoD Baseline.}
We compare our approach with an adapted version of the Chain-of-Diagnosis (CoD) framework \citep{chen2025cod}. The CoD is a multi-turn diagnostic pipeline that retrieves candidate diseases via embedding similarity between patient-mentioned symptoms and disease-associated symptoms, and finalizes the diagnosis once a confidence score over the top-3 candidates exceeds a fixed threshold.

CoD is a suitable baseline because it shares the same high-level structure as our system: both methods (i) identify a small set of plausible disease candidates and (ii) use external medical knowledge to either ask CQs or output a diagnosis. However, they differ in how knowledge is represented and how the stopping decision is made. During CQ generation, CoD provides the model with disease-centric textual evidence (e.g., disease--symptom/treatment lists) while our method provides the model with subgraph expanded from diagnostic hypotheses. Furthermore, CoD generates a confidence score at each turn and forces termination once this score exceeds a predefined threshold. In our system, by contrast, the model reasons over the dialogue history and the subgraph to decide whether to continue asking CQs or to produce a final diagnosis.

To ensure that both CoD and our framework are grounded in the same underlying medical knowledge, we do not use CoD’s original disease database. Instead, for each candidate disease, we construct CoD’s retrieval context from our $\mathcal{G}$ by extracting the disease node and its 1-hop connected attributes (e.g., \texttt{symptom}, \texttt{cause}, and \texttt{risk\_factor}) and providing this information in CoD’s required textual format. 

\paragraph{Use of HG in KG-Based Settings.}
Unless otherwise specified, all settings use the same HG as our proposed system, a supervised Qwen-2.5-7B-Instruct model with a classification head trained for multi-label disease prediction. For experiments with general-purpose models, HG first produces disease hypotheses and the corresponding hypothesis-driven subgraph, and the general-purpose model is then used as the HV to generate CQs or final diagnoses. This design isolates the effect of the HV under the same HG setup. The adapted CoD baseline is the only exception, as it follows its own similarity-based candidate retrieval and thresholding procedure.

\subsection{Experimental Setup}
\paragraph{Evaluation Metrics.}
We evaluate diagnostic accuracy using Recall@$k$ ($k=1,2,3,4$), defined as the average proportion of ground-truth diagnoses ($d_{gold}$) included within the top-$k$ predicted diagnoses across all dialogues. We also measure diagnostic efficiency by the average number of dialogue turns. Because effective history-taking prioritizes a small set of highly informative questions rather than exhaustive inquiry, lower turn counts with comparable accuracy indicate a more practical diagnostic process.

\paragraph{Evaluation Set Details.}
All results are reported on a held-out test set of 275 patient profiles. Each profile contains up to two ground-truth diseases (252 single-disease cases and 23 double-disease cases). As a result, Recall@1 is theoretically upper-bounded at 0.958, whereas the maximum Recall@$k$ for $k \ge 2$ is 1.000.

\paragraph{Maximum Turn Limit.}
To avoid infinite loops, we set a maximum of 50 dialogue turns; sessions exceeding this limit are terminated and counted as failures. This threshold is based on an empirical analysis showing that 91\% of patient profiles contain fewer than 50 distinct information units. Since these profiles originate from emergency department records and often include more detail than required for outpatient initial consultations, informing all recorded symptoms is unnecessary and would be prohibitively burdensome to real patients. We therefore treat 50 turns as sufficient and practical for dialogue-based history-taking.

\section{Results}
\subsection{Diagnostic Performance}

\paragraph{Impact of the KG and the HG.}
To analyze the impact of $\mathcal{G}$ and HG, we conduct an ablation study using GPT-4.1-mini across three settings, as shown in Table~\ref{tab:ablation}. Baseline (row 1) relying solely on parametric knowledge achieves Recall@1 of 0.142, indicating limited diagnostic capability. The KG-based variant (row 2) operates without the HG by performing a 2-hop expansion starting from non-disease attribute nodes semantically similar to the patient utterance. The improved Recall@1 of 0.218 demonstrates the benefit of KG grounding. Our full framework (row 3) incorporates the HG to estimate disease probabilities and dynamically filter the extracted subgraph. It achieves the best performance across all metrics (Recall@1--4), increasing Recall@1 to 0.243. These results demonstrate both higher diagnostic accuracy and more efficient diagnosis.

\begin{table}[t]
    \centering
    \resizebox{\columnwidth}{!}{%
        \begin{tabular}{lccccc}
            \toprule
             & \multicolumn{4}{c}{\textbf{Recall@$k \uparrow$}} & \\
            \cmidrule(lr){2-5}
            \textbf{Models} & \textbf{1} & \textbf{2} & \textbf{3} & \textbf{4} & \textbf{Turns $\downarrow$} \\
            \midrule
            GPT-4.1-mini & 0.142 & 0.218 & 0.218 & 0.218 & \textbf{7.5} \\
            \quad + KG & \underline{0.218} & \underline{0.319} & \underline{0.329} & \underline{0.343} & 15.3 \\
            \quad + KG + HG & \textbf{0.243} & \textbf{0.357} & \textbf{0.376} & \textbf{0.383} & \underline{14.1} \\
            \bottomrule
        \end{tabular}%
    }
    \caption{Ablation study on the impact of the diagnostic knowledge graph and the Hypothesis Generator (HG) on diagnostic performance. Best scores are in \textbf{bold}, and second-best scores are \underline{underlined}. The `$+$~HG' row uses the trained Qwen-2.5-7B-Instruct $+$ classification head (\S\ref{sec:training_details}); the diagnosis backbone in all rows is GPT-4.1-mini.}
    \label{tab:ablation}
\end{table}

\paragraph{Comparison with Baseline Models.}
Table~\ref{tab:model_performance} compares our SFT model with baselines that incorporate the same HG and $\mathcal{G}$ (except for CoD). 

Notably, our SFT model outperforms substantially larger frontier models. It consistently surpasses GPT-4.1-mini across all $k$ values (Recall@1--4). While Claude-3.5-Sonnet achieves the highest Recall@1, it is significantly larger and our model still outperforms it for the remaining metrics (Recall@2--4 and Turns). 

Our SFT model significantly outperforms its base model, Qwen2.5-7B-Instruct (Base), across all $k$ values (Recall@1--4), while reducing the average dialogue turns from 12.4 to 6.9. This demonstrates that fine-tuning on our synthetic diagnostic dialogues enables the model to leverage these components more effectively. The adapted CoD baseline with its optimal confidence threshold terminates too quickly and exhibits lower recall, indicating premature diagnosis. The detailed evaluation results across multiple confidence thresholds are provided in Appendix~\ref{app:cod_baseline}. 

\paragraph{Error Analysis.}
We conducted a qualitative analysis to understand the misdiagnosed cases. We identified four primary error patterns: (1) the discrepancy between textbook and real-world patient narratives, (2) the difficulty of diagnosing solely through conversation, (3) confusion between acute and chronic conditions, and (4) limitations arising from label granularity in clinical data. Details are provided in Appendix~\ref{app:Error_Analysis}.

\begin{table}[t]
    \centering
    \resizebox{\columnwidth}{!}{%
        \begin{tabular}{lccccc} 
            \toprule
             & \multicolumn{4}{c}{\textbf{Recall@$k \uparrow$}} & \\
            \cmidrule(lr){2-5}
            \textbf{Models} & \textbf{1} & \textbf{2} & \textbf{3} & \textbf{4} & \textbf{Turns $\downarrow$} \\
            \midrule
            
            \multicolumn{6}{l}{\textit{\textbf{Large Models}}} \\
            \midrule
            GPT-4.1-mini         & 0.243 & \underline{0.357} & \underline{0.376} & \underline{0.383} & 14.1 \\
            Claude-3.5-Sonnet    & \textbf{0.267} & 0.329 & 0.345 & 0.345 & 9.4 \\
            Gemini-2.0-Flash     & 0.187 & 0.241 & 0.250 & 0.251 & 37.0 \\
            Llama-3.3-70B & 0.186 & 0.283 & 0.306 & 0.313 & 13.9 \\
            Qwen-2.5-72B & 0.201 & 0.255 & 0.260 & 0.260 & 35.2 \\
            \midrule
            
            \multicolumn{6}{l}{\textit{\textbf{Small Models (7B)}}} \\
            \midrule
            CoD                  & 0.106 & 0.149 & 0.196 & 0.207 & \textbf{2.0} \\
            Qwen-2.5 (Base)  & 0.173 & 0.218 & 0.218 & 0.222 & 12.4  \\
            \textbf{SFT (\textit{Ours})} & \underline{0.250} & \textbf{0.361} & \textbf{0.394} & \textbf{0.418} & \underline{6.9}   \\
            \bottomrule
    \end{tabular}
    }
    \caption{Diagnostic performance of different models on Recall and the number of Turns. Best scores are in \textbf{bold}, and second-best scores are \underline{underlined}. All models share the same trained HG and graph $\mathcal{G}$ except CoD; only the diagnosis backbone (the HV role) differs. CoD uses its own retrieval-based candidate selection over the same $\mathcal{G}$ (\S\ref{subsec:baselines}).}
    \label{tab:model_performance}
\end{table}

\subsection{Analysis of Hypothesis Generator}

\paragraph{Validation of HG Architecture.}
To validate our architectural choice for the HG, we compare it with three alternatives: (1) a zero-shot generative baseline using GPT-4.1-mini (Gen.); (2) an embedding-based retrieval baseline using SapBERT \citep{liu-etal-2021-self} (Retriever), which retrieves disease nodes based on the embedding similarity between patient-reported symptoms and disease symptoms; and (3) a fine-tuned generative baseline based on Qwen-2.5-7B-Instruct (Gen., FT) that directly generates disease names. Recall@$k$ is computed on truncated dialogues from the HG test split (see \S~\ref{sec:training_details}).

\begin{figure}[t]
    \centering
    \includegraphics[width=\columnwidth]{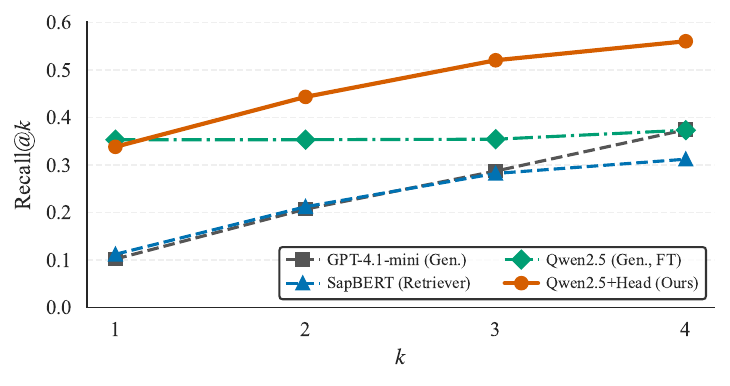}
    \caption{Comparison of disease hypothesis generation performance (Recall@$k$) across different methods.}
    \label{fig:recall_graph}
\end{figure}

As shown in Figure~\ref{fig:recall_graph}, GPT-4.1-mini (Gen.) and SapBERT (Retriever) show limited coverage, suggesting that zero-shot generation or embedding similarity alone is insufficient for handling incomplete multi-turn histories. While the fine-tuned generative baseline, Qwen2.5 (Gen., FT), performs strongly at $k{=}1$, its improvement plateaus as $k$ increases, indicating a failure to generate reliable candidates for the 2nd to 4th ranks. In contrast, Qwen2.5 + Head (\textit{Ours}) achieves significantly higher coverage at larger $k$, reaching a Recall@4 of 0.560. Since subgraph construction incorporates multiple anchor nodes rather than a single selection, high Recall@$k$ is more critical than Recall@1 to ensure the ground-truth disease is captured within the extracted subgraph. This demonstrates that our classification head provides more reliable ranking, establishing the HG as a robust engine for hypothesis generation.

\begin{table*}[t]
    \centering
    \footnotesize
    \resizebox{\linewidth}{!}{%
        \begin{tabular}{lcccccccccc}
            \toprule
             & \multicolumn{4}{c}{\textbf{Diagnosis Recall@$k \uparrow$}} & & \multicolumn{4}{c}{\textbf{HG Recall@$k \uparrow$}} & \\
            \cmidrule(lr){2-5} \cmidrule(lr){7-10}
            \textbf{HG Setting} & \textbf{1} & \textbf{2} & \textbf{3} & \textbf{4} & \textbf{Turns $\downarrow$} & \textbf{1} & \textbf{2} & \textbf{3} & \textbf{4} & \textbf{Sub Recall $\uparrow$} \\
            \midrule
            $n{=}1$, $\tau{=}0.003$ & 0.204 & \underline{0.347} & \underline{0.376} & \underline{0.411} & \underline{7.1} & 0.190 & - & - & - & \underline{0.538} \\
            $n{=}2$, $\tau{=}0.005$ & \textbf{0.250} & \textbf{0.361} & \textbf{0.394} & \textbf{0.418} &  \textbf{6.9} & 0.142 & 0.256 & - & - & \textbf{0.559} \\
            $n{=}3$, $\tau{=}0.01$  & 0.218 & 0.305 & 0.361 & 0.387 & 7.2 & 0.197 & 0.270 & 0.317 & - & 0.482 \\
            $n{=}4$, $\tau{=}0.015$ & \underline{0.229} & 0.329 & 0.368 & \textbf{0.418} & 7.2 & 0.185 & 0.277 & 0.314 & 0.343 & 0.423 \\
            \bottomrule
        \end{tabular}%
    }
    \caption{Diagnostic performance of different hyperparameters on 3-hop subgraph expansion setting. Best scores are in \textbf{bold}, and second-best scores are \underline{underlined}.}
    \label{tab:HG_performance}
\end{table*}

\paragraph{Impact of the HG Hyperparameters.}
Table~\ref{tab:HG_performance} presents the impact of HG hyperparameters on the end-to-end diagnostic performance of our system. Here, $n$ denotes the number of top diseases selected by the HG, and $\tau$ denotes the probability threshold used to filter low-confidence disease nodes during subgraph extraction. We calibrated these parameters to ensure that the average number of disease nodes in the extracted subgraphs remained comparable across settings. We report three complementary metrics: (i) \textit{Diagnosis Recall@$k$}, which measures whether the ground-truth disease is included in the top-$k$ final diagnoses; (ii) \textit{HG Recall@$k$}, the proportion of ground-truth diseases captured within the top-$k$ anchor nodes directly predicted by the HG; and (iii) \textit{Subgraph Recall (Sub Recall)}, which indicates whether the ground-truth disease is contained within the 3-hop subgraph expanded from those anchor nodes. All metrics are computed at the final turn of each dialogue.

The configuration with $n{=}2$ and $\tau{=}0.005$ yields the best performance (\textit{Diagnosis Recall@1} of 0.250 and \textit{@4} of 0.418), coinciding with the highest \textit{Sub Recall} of 0.559. Given the controlled subgraph size, this success suggests that concentrating on a few high-confidence anchors ($n{=}2$) while allowing broad exploration of their neighbors via a lenient threshold ($\tau{=}0.005$) is more effective for hypothesis verification than distributing attention across a broader set of anchors (i.e., $n{=}4$, where \textit{Sub Recall} drops to 0.423). This shows that accurately containing the ground-truth disease leads to effective hypothesis verification.

\paragraph{Performance of the Standalone HG.}
Given that the HG already produces a ranked list of disease candidates, one might question the necessity of the subsequent graph-based reasoning step. To address this, we compare the accuracy of the HG's hypotheses against the final diagnosis. Table~\ref{tab:HG_performance} demonstrates that the full pipeline significantly improves upon the HG's own ranking. In our best configuration ($n{=}2, \tau{=}0.005$), the \textit{Diagnosis Recall@1} (0.250) substantially surpasses the \textit{HG Recall@1} (0.142). This gap indicates that the HG serves primarily as a preliminary filter, while reasoning grounded in the extracted subgraph is essential for correcting false positives and refining the diagnostic ranking.

\subsection{Evaluation of Patient Simulator}
\label{sec:patient_simulator_eval}
To validate if our specificity-augmented simulator better reflects real-world patients than the original PatientSim, we conducted an expert evaluation.

\paragraph{Experimental Setup.} 
We recruited three board-certified physicians to assess the realism of the patient responses. The evaluation set consisted of 41 distinct dialogues randomly sampled from dialogues generated by our system during inference time. For each scenario, the dialogue history was generated by the baseline PatientSim, and only the last patient response differed between the two simulators. 
This design evaluates both responses under the same dialogue context, isolating the effect of the final response. It is also conservative for our simulator, since the specificity-augmented response is judged within a dialogue history generated by the baseline simulator. The study was blinded: responses were shown as `Response~A' and `Response~B' in randomized order with no system identifier. The evaluation interface is provided in Appendix~\ref{app:patient_simulator_evaluation_details}.

\paragraph{Results.} 
Across 41 dialogue scenarios, physicians preferred our simulator in $46.3\%$ of cases, compared to $24.4\%$ for the baseline, with $29.3\%$ rated as comparable. Despite the inherent subjectivity of evaluating patient realism, two out of three physicians reached a consensus in $70\%$ of the cases. Among these cases, our simulator was preferred in $75\%$ of the cases. This preference emerged even though the preceding dialogue context was generated by the baseline PatientSim. 

We further analyze inter-rater agreement. Fleiss' $\kappa$ is near zero, partly because the preference labels are skewed, so we also report the prevalence-robust Gwet's AC1. AC1 is $0.06$ overall but rises to $0.48$ on cases where evaluators made definitive judgments within their area of confidence, suggesting that the low agreement reflects specialty-limited uncertainty rather than a fundamental conflict over simulator realism. The full analysis is provided in Appendix~\ref{app:patient_simulator_agreement}.

Qualitative feedback further supports this result. Physicians noted that vagueness in patient responses was a key factor in making the dialogue feel clinically authentic, aligning with our goal of low-specificity symptom reporting. In a complementary review of full 10 inference dialogues, two board-certified physicians found the modeled vagueness clinically plausible, while noting occasional artifacts such as abrupt late symptom disclosure and patients volunteering their own differential diagnosis (e.g., ``it seems related to my previous clot problem''). Detailed feedback is provided in Appendix~\ref{app:patient_simulator_comment}.


\subsection{Evaluation of CQ Quality}
\label{sec:cq_eval}
To evaluate the quality of CQs generated by the Hypothesis Verifier (HV), we conducted an interactive human evaluation by medical experts.

\paragraph{Experimental Setup.} 
We recruited four board-certified physicians to participate in interactive diagnostic sessions. Each physician role-played as a patient using 5 unique profiles from the evaluation set, totaling 20 independent sessions. The physicians evaluated CQs on a 5-point Likert scale across three dimensions: \textit{Essentiality} (necessity for diagnosis), \textit{Conversational Flow} (logical progression), and \textit{Clinical Authenticity} (alignment with real-world history-taking), while also providing qualitative feedback to support their evaluations. 

\paragraph{Results.} 
Our system scored high in \textit{Clinical Authenticity} (4.25), confirming that CQs closely mirror clinician inquiry patterns. \textit{Essentiality} (3.5) and \textit{Conversational Flow} (3.45) also received positive ratings, indicating moderate efficiency in information gathering. Qualitative feedback highlighted that the overall process of history-taking was clinically natural and appropriate. Furthermore, experts noted that the CQs were strategically targeted toward refining the diagnostic hypotheses identified by HG, even when initial patient descriptions were vague. At the same time, physicians noted that some critical cues (e.g., red-flag symptoms or past medical history) were occasionally underexplored, and that a few candidate diagnoses were less clinically plausible, indicating room for improvement in question prioritization. Detailed evaluation protocols, metrics and qualitative feedback are provided in Appendix \ref{app:cq_eval_details}.


\subsection{Robustness to Patient Persona Variations}
We analyze diagnostic performance across four persona attributes: language proficiency, personality, recall level, and confusion level. We compare three configurations: (i) a baseline relying solely on internal knowledge (GPT-only), (ii) a zero-shot GPT grounded in HG and KG (GPT+KG), and (iii) our fully fine-tuned model grounded in HG and KG (SFT+KG). GPT and SFT denote GPT-4.1-mini and our fine-tuned model (Qwen-2.5-7B-Instruct) respectively.

\begin{figure}[t!]
    \centering
    \includegraphics[width=\linewidth]{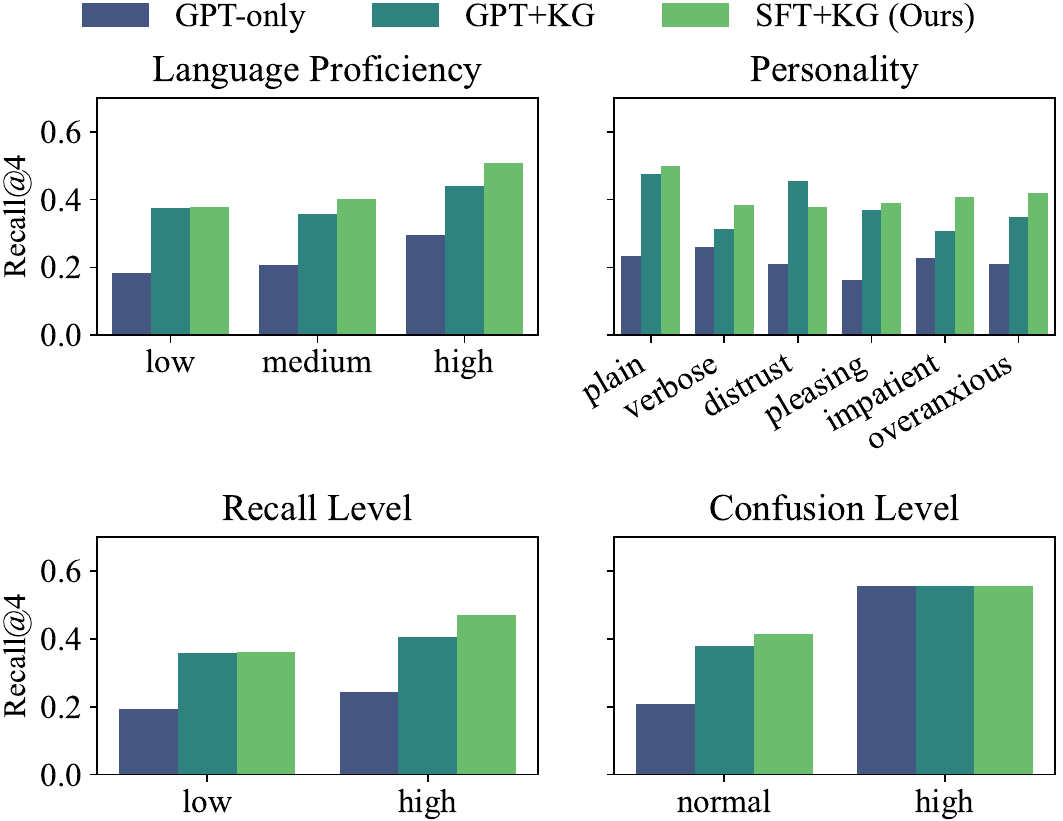}
     \caption{Robustness analysis under diverse patient personas (Recall@4).}
    \label{fig:robustness}
\end{figure}

As shown in Figure~\ref{fig:robustness}, KG grounding consistently improves Recall@4 over the GPT-only baseline across all persona variations. Furthermore, the performance gap between our SFT+KG and GPT-only becomes more pronounced as language proficiency decreases. SFT+KG remains competitive across all personality types, including emotionally challenging personas such as \textit{distrust} and \textit{overanxious}. Overall, these results suggest that integrating the KG ensures robust diagnosis even for challenging patient profiles, effectively handling linguistic barriers, emotionally diverse behaviors, and varying memory capabilities. Note that unlike other attributes, the high \textit{confusion} level has a limited sample size, reducing the reliability of the corresponding results.

\begin{table*}[t]
\centering
\small
\setlength{\tabcolsep}{4pt}
\begin{tabular}{lcccccccccc}
\toprule
& \multicolumn{4}{c}{OOG Recall@$k$} & \multicolumn{4}{c}{IG Recall@$k$} & & \\
\cmidrule(lr){2-5}\cmidrule(lr){6-9}
Setting & 1 & 2 & 3 & 4 & 1 & 2 & 3 & 4 & all@$4$ & Turns \\
\midrule
Original system & 0.000 & 0.000 & 0.000 & 0.000 & 0.250 & 0.361 & 0.394 & 0.418 & 0.308 & 6.9 \\
\midrule
``Other'' node  & 0.724 & 0.745 & 0.786 & 0.898 & 0.122 & 0.255 & 0.298 & 0.313 & 0.466 & 7.4 \\
KG augmentation & 0.102 & 0.153 & 0.173 & 0.194 & 0.200 & 0.305 & 0.347 & 0.414 & 0.357 & 7.6 \\
\bottomrule
\end{tabular}
\caption{Diagnostic performance of OOG settings and original setting. original setting uses the 338-disease graph, while the two OOG strategies either add an 'Other' node or augment the KG with previously OOG diseases. OOG Recall@$k$ measures correct OOG handling: answering 'Other' for the 'Other' node strategy, or naming the now in-graph disease for KG augmentation. IG Recall@$k$ measures performance on previously in-graph diseases. The overall column (all@$4$) is computed under the real IG:OOG prevalence ratio ($2.81{:}1$).}
\label{tab:ood_summary}
\end{table*}

\subsection{Handling Out-of-Graph Diseases}
\label{sec:ood}

A natural concern is how the system behaves when the ground-truth disease lies outside $\mathcal{G}$: because the HG ranks only in-graph diseases, an out-of-graph (OOG) ground truth is a guaranteed false negative. From the 5{,}420 OOG profiles identified during preprocessing (\S\ref{subsubsec:data_preprocessing}), we build held-out OOG evaluation sets (108 validation / 98 test) and study two complementary strategies. 
(1) We add an 'Other' node and train the HG over 338 diseases plus the 'Other' label, so that the system can flag an OOG case by answering 'Other' rather than forcing an in-graph diagnosis. 
(2) We augment the KG by expanding $\mathcal{G}$ to 528 diseases. For the previously OOG diseases, we add those diseases as new \texttt{disease} nodes and construct their associated \texttt{symptom}, \texttt{cause}, and \texttt{risk\_factor} attributes from the patient profiles. This turns previously OOG diseases into in-graph nodes, making them directly diagnosable by the same framework.

Table~\ref{tab:ood_summary} compares the original system, with the two OOG-handling strategies. The original system has no mechanism for representing diseases outside the 338-disease graph and therefore cannot recover any OOG case, yielding OOG Recall@$k$ of $0.000$ for all $k$.
The 'Other' node is effective at flagging OOG cases. By learning to answer ``Other'' rather than forcing an in-graph diagnosis, it reaches test OOG Recall@4 of $0.898$. This comes at some cost to in-graph performance.
KG augmentation addresses a harder setting: instead of merely flagging an OOG case, the system must name the specific disease after adding it to the graph. This raises test OOG Recall@4 from $0.000$ to $0.194$ while preserving most in-graph performance (IG Recall@4 $=0.414$). 
Thus, the two strategies are complementary: the ``Other'' node supports safe flagging, whereas KG augmentation expands the set of diseases that can be diagnosed directly. Appendix~\ref{app:graph_expansion} reports the full validation sweep over the training set's IG/OOG ratio.

Two further analyses put the augmentation result in context. First, the HG recognizes the added OOG diseases far more often than the end-to-end score suggests: its HG-level OOG Recall@4 rises from $0.000$ to $0.53$ across the sweep, so most of the residual gap lies in downstream verification rather than in recognizing that a disease is out-of-graph. Second, because the ``Other'' node only needs to flag a case whereas augmentation must name the exact disease, a criterion-matched comparison---crediting augmentation whenever its top-4 contains any out-of-graph disease, as the ``Other'' node is---raises its OOG Recall@4 from $0.194$ to $0.541$, roughly halving the gap to the ``Other'' node.

\section{Conclusion}
In this work, we presented a conversational diagnosis system designed to support clinical history-taking under incomplete and underspecified patient information. Motivated by how clinicians iteratively refine diagnostic hypotheses through targeted questioning, our system models conversational diagnosis as a two-step process of hypothesis generation and hypothesis verification, grounded in a diagnostic knowledge graph. 

Experimental results show that graph grounding substantially improves both diagnostic accuracy and efficiency compared to baselines that rely on parametric knowledge or similarity-based retrieval. In addition, our hypothesis-driven subgraph construction and graph-grounded training strategy improve robustness under diverse patient behaviors. Overall, these findings highlight the potential of combining explicit diagnostic knowledge with interactive language models for reliable and interpretable conversational diagnosis.

\section{Limitations}
Our work has limitations inherent to its experimental scope. 
First, we used a patient simulator rather than real-world clinical encounters. This was mainly for ethical considerations regarding testing experimental AI with vulnerable patients, and we instead strengthened our evaluation with patient role-plays performed by physicians. 
Second, the diagnostic scope is bounded by the current coverage of the diagnostic knowledge graph, which excludes some conditions (including rare diseases). However, this limitation is not fundamental. As we demonstrate through a graph-expansion experiment, the framework can be adopted as-is on an expanded graph, while preserving in-graph diagnostic accuracy.
Third, following PatientSim, our simulator does not condition each response on the full dialogue history. Incorporating dialogue history is a promising direction for improving long-range consistency in simulated patient behavior.
Fourth, even with KG expansion, some diseases remain difficult to diagnose from text-only history-taking alone. Many require imaging, laboratory tests, or procedure. In such cases, the 'Other' node provides a safer behavior by flagging uncertainty rather than forcing an in-graph diagnosis.


\bibliography{main}
\bibliographystyle{acl_natbib}

\clearpage
\section*{Appendix}
\appendix

\setlength{\textfloatsep}{6pt}   
\setlength{\floatsep}{6pt}       
\setlength{\intextsep}{6pt}      

\section{Details of Diagnostic Schemas and Diagnostic Knowledge Graph}
\label{app:kg_details}

\subsection{Schema Source and Graph Construction}
The schemas are drawn from \textit{10 Minutes CPX}\footnote{\url{https://www.yes24.com/Product/Goods/102918444}}, a standard Korean textbook for the Clinical Performance Examination (CPX) of the Korean Medical Licensing Examination. It covers 48 schemas across major clinical domains, including cardiopulmonary complaints (e.g., cough, chest pain), gastrointestinal complaints (e.g., abdominal pain, hematemesis, diarrhea), neuropsychiatric complaints (e.g., headache, dizziness), musculoskeletal/dermatologic/endocrine complaints (e.g., back pain, rash), genitourinary/obstetric complaints (e.g., dysuria, hematuria), and general systemic symptoms (e.g., weight loss, fever). We exclude the counseling-focused section (e.g., smoking cessation, substance abuse) as misaligned with our outpatient initial-consultation setting, and use only the diagnostic history-taking schemas rather than the accompanying examination, treatment, and example-case materials.

Each schema provides (i) a differential diagnosis list for the symptom group and (ii) discriminative history-taking cues (key symptom attributes, associated findings, and risk factors). We translated the original Korean schemas into English and reformatted them into a unified graph, extracting disease nodes and their clinical attributes.

The four node types (\texttt{disease}, \texttt{symptom}, \texttt{cause}, \texttt{risk\_factor}) are linked by three relations: \texttt{caused\_by} connects a \texttt{symptom} to a \texttt{disease} it clinically manifests; \texttt{can\_cause} links a \texttt{cause} to a \texttt{disease} it can etiologically give rise to; and \texttt{is\_a\_risk\_factor\_of} connects a \texttt{risk\_factor} to a \texttt{disease} whose likelihood it increases. Figure~\ref{fig:kg_schema} shows the graph schema, and the \textbf{Paper}~($\mathcal{G}$) column of Table~\ref{tab:kg_expansion} reports its node- and edge-level statistics.

\begin{center}
\begin{minipage}{0.8\columnwidth}
    \centering
    \includegraphics[width=0.8\columnwidth]{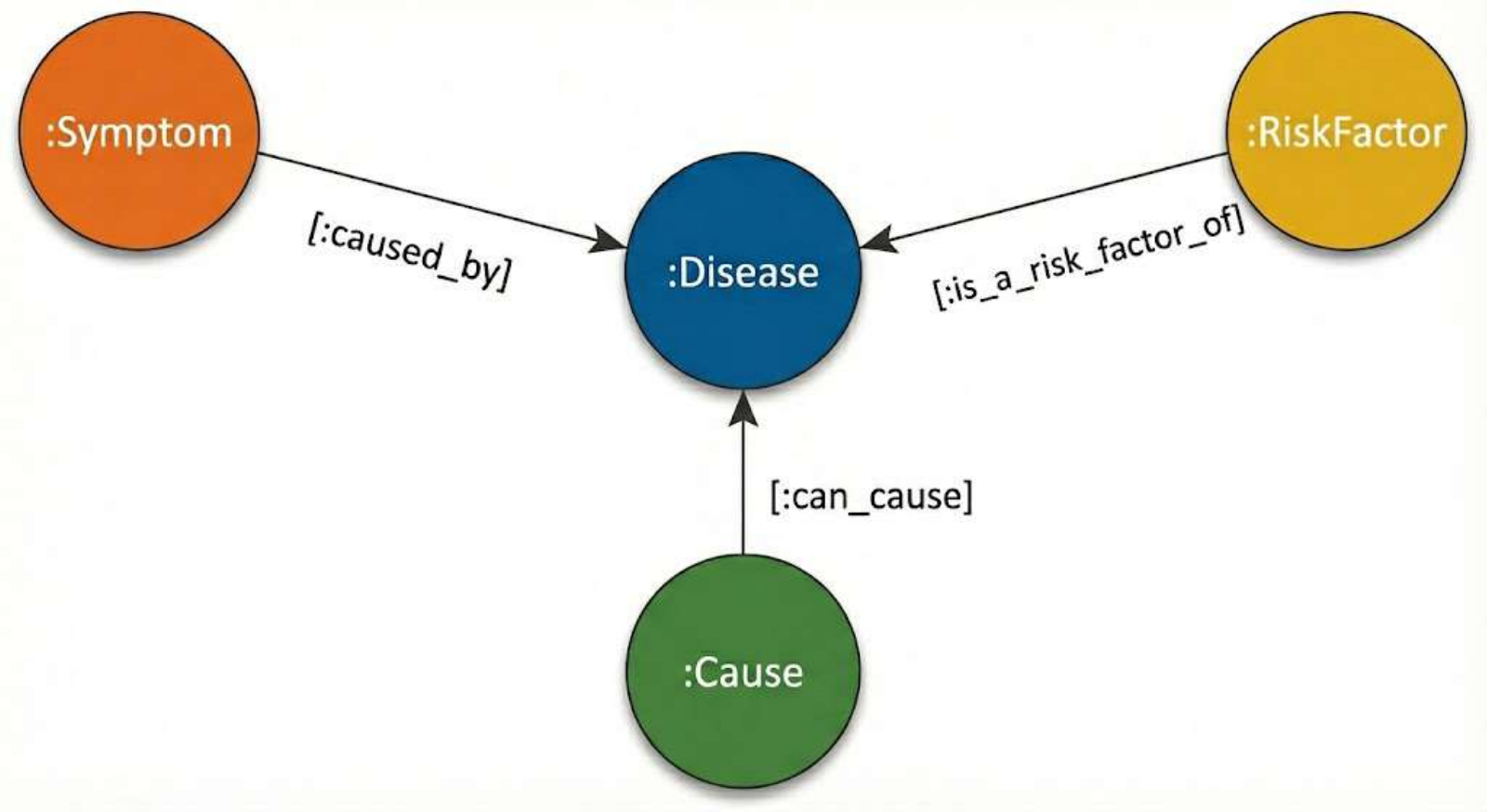}
    \captionof{figure}{Schema of the diagnostic knowledge graph.}
    \label{fig:kg_schema}
\end{minipage}
\end{center}

\subsection{Graph Expansion and Out-of-Graph Evaluation}
\label{app:graph_expansion}

\subsubsection{Expanded Graph, Data Splits, and Soft Coverage}
We expand $\mathcal{G}$ by mining diagnostic schemas (Claude-Sonnet) for the previously out-of-graph diseases, adding their disease and attribute nodes (Table~\ref{tab:kg_expansion}), and re-train and re-run the identical pipeline on the expanded $\mathcal{G}^{+}$ with no architectural change.

\begin{table}[t]
\centering
\scriptsize
\setlength{\tabcolsep}{4pt}
\resizebox{\columnwidth}{!}{%
\begin{tabular}{lrr}
\toprule
 & \textbf{Paper} ($\mathcal{G}$) & \textbf{Augmented} ($\mathcal{G}^{+}$) \\
\midrule
\multicolumn{3}{l}{\emph{Nodes}} \\
\quad \texttt{disease}      & 338 & 528 \\
\quad \texttt{symptom}      & 847 & 1{,}167 \\
\quad \texttt{cause}        & 266 & 594 \\
\quad \texttt{risk\_factor} & 282 & 457 \\
\quad Total                 & \textbf{1{,}733} & \textbf{2{,}746} \\
\midrule
\multicolumn{3}{l}{\emph{Edges}} \\
\quad \texttt{caused\_by}             & 2{,}630 & 4{,}295 \\
\quad \texttt{can\_cause}             & 805 & 1{,}807 \\
\quad \texttt{is\_a\_risk\_factor\_of} & 500 & 1{,}371 \\
\quad Total                          & \textbf{3{,}935} & \textbf{7{,}473} \\
\bottomrule
\end{tabular}}
\caption{Node/edge statistics of the original vs.\ expanded graph. The expansion adds 190 diseases; the original is strictly contained in the expanded one.}
\label{tab:kg_expansion}
\end{table}

OOG profiles are drawn from the 5{,}420 pool and added to the original IG profiles; validation and test sets are fixed while only the training OOG count varies with $r$ (Table~\ref{tab:ood_split_app}).

\begin{table}[t]
\centering
\small
\resizebox{\columnwidth}{!}{%
\begin{tabular}{llcccccccc}
\toprule
 & Split & $0\%$ & $5\%$ & $10\%$ & $15\%$ & $20\%$ & $25\%$ & $30\%$ & $35\%$ \\
\midrule
\multirow{4}{*}{HG} & train IG  & 1{,}371 & 1{,}371 & 1{,}371 & 1{,}371 & 1{,}371 & 1{,}371 & 1{,}371 & 1{,}371 \\
                    & train OOG & 0 & 72 & 152 & 242 & 343 & 457 & 588 & 738 \\
                    & valid     & \multicolumn{8}{c}{IG 171 \;+\; OOG 34} \\
                    & test      & \multicolumn{8}{c}{IG 170 \;+\; OOG 73} \\
\midrule
\multirow{4}{*}{HV} & train IG  & 1{,}439 & 1{,}439 & 1{,}439 & 1{,}439 & 1{,}439 & 1{,}439 & 1{,}439 & 1{,}439 \\
                    & train OOG & 0 & 76 & 160 & 254 & 360 & 480 & 617 & 775 \\
                    & valid     & \multicolumn{8}{c}{IG 304 \;+\; OOG 108} \\
                    & test      & \multicolumn{8}{c}{IG 275 \;+\; OOG 98} \\
\bottomrule
\end{tabular}}
\caption{IG/OOG data splits across training OOG ratios. IG profiles reuse the original experiment; OOG profiles are added. Validation/test are fixed; only the training OOG count varies with $r$.}
\label{tab:ood_split_app}
\end{table}

As a preliminary check, for each of the 98 OOG test profiles we find its \emph{closest in-graph relative}---the in-graph disease with the highest Jaccard attribute overlap on $\mathcal{G}^{+}$---and ask how often the original 338-disease HG nominates it. The relatives are clinically coherent (e.g., \emph{deep vein thrombosis}$\to$\emph{pulmonary embolism}, \emph{ulcerative colitis}$\to$\emph{Crohn's disease}), confirming that OOG cases share attributes with in-graph neighbors. Yet the HG nominates the relative only 13\% of the time at top-4 (21\% at top-10), so attribute sharing gives only weak soft coverage and most OOG cases remain hard false negatives---motivating the two strategies of \S\ref{sec:ood}.

\subsubsection{End-to-End OOG Results: Full Training-Ratio Sweep}
\label{app:ood_sweep}
Table~\ref{tab:ood_sweep_full} gives the full sweep for both strategies, with the test row at the validation-selected ratio. For the \textbf{``Other'' node}, validation OOG Recall@4 climbs to $0.944$; since larger $r$ trades IG accuracy for OOG recovery, we select the best-balancing ratio ($25\%$), giving test OOG Recall@4 $0.898$. For \textbf{KG augmentation}, the HG recognizes the OOG diseases (HG-level OOG Recall@4 up to $0.53$), but on the raw LLM-mined graph this did not reach end-to-end diagnosis ($\sim$0.10): the added neighborhoods were sparse and the HV had been trained on gold-anchored ``oracle'' subgraphs absent at inference. Rebuilding the attributes from MIMIC-validated evidence and re-training the HV on inference-matched subgraphs raises end-to-end OOG Recall@4 to $0.227$ (validation) / $0.194$ (test) while preserving IG accuracy. The residual gap to the ``Other'' node (criterion-matched comparison in \S\ref{sec:ood}) reflects a modality ceiling: many OOG diseases require imaging, laboratory, or procedural confirmation that a text-only history cannot surface (e.g., deep vein thrombosis, bacteremia, complete atrioventricular block, ulcerative colitis).

\begin{table*}[t]
\centering
\scriptsize
\setlength{\tabcolsep}{4pt}
\begin{tabular}{lcccccccccc}
\toprule
& \multicolumn{4}{c}{OOG Recall@$k$} & \multicolumn{4}{c}{IG Recall@$k$} & & \\
\cmidrule(lr){2-5}\cmidrule(lr){6-9}
OOG ratio & 1 & 2 & 3 & 4 & 1 & 2 & 3 & 4 & all@$4$ & Turns \\
\midrule
\multicolumn{11}{l}{\emph{Original system (no out-of-graph handling)}} \\
$0\%$ & 0.000 & 0.000 & 0.000 & 0.000 & 0.250 & 0.361 & 0.394 & 0.418 & 0.308 & 6.9 \\
\midrule
\multicolumn{11}{l}{\emph{Strategy~(1) ``Other'' node --- validation sweep}} \\
$0\%$ & 0.000 & 0.000 & 0.009 & 0.204 & 0.190 & 0.254 & 0.316 & 0.359 & 0.318 & 6.6 \\
$5\%$ & 0.000 & 0.028 & 0.111 & 0.491 & 0.203 & 0.260 & 0.293 & 0.326 & 0.369 & 6.9 \\
$10\%$ & 0.324 & 0.343 & 0.463 & 0.778 & 0.190 & 0.269 & 0.308 & 0.334 & 0.451 & 6.4 \\
$15\%$ & 0.380 & 0.398 & 0.491 & 0.713 & 0.164 & 0.255 & 0.315 & 0.351 & 0.446 & 6.9 \\
$20\%$ & 0.620 & 0.639 & 0.676 & 0.833 & 0.129 & 0.225 & 0.272 & 0.294 & 0.435 & 7.2 \\
\textbf{25\%} & 0.648 & 0.648 & 0.676 & 0.870 & 0.143 & 0.259 & 0.302 & 0.330 & \textbf{0.472} & 7.2 \\
$30\%$ & 0.806 & 0.815 & 0.824 & 0.917 & 0.098 & 0.195 & 0.253 & 0.294 & 0.458 & 7.4 \\
$35\%$ & 0.833 & 0.833 & 0.852 & 0.944 & 0.093 & 0.210 & 0.258 & 0.283 & 0.457 & 7.6 \\
\quad\emph{test} ($25\%$) & 0.724 & 0.745 & 0.786 & 0.898 & 0.122 & 0.255 & 0.298 & 0.313 & 0.466 & 7.4 \\
\midrule
\multicolumn{11}{l}{\emph{Strategy~(2) KG augmentation --- validation sweep}} \\
$0\%$ & 0.019 & 0.056 & 0.065 & 0.065 & 0.269 & 0.351 & 0.384 & 0.442 & 0.343 & 7.5 \\
$5\%$ & 0.037 & 0.065 & 0.065 & 0.065 & 0.284 & 0.353 & 0.401 & 0.451 & 0.350 & 7.7 \\
$10\%$ & 0.056 & 0.093 & 0.120 & 0.120 & 0.282 & 0.370 & 0.416 & 0.478 & 0.384 & 8.0 \\
$15\%$ & 0.065 & 0.111 & 0.130 & 0.148 & 0.270 & 0.340 & 0.397 & 0.450 & 0.371 & 8.1 \\
$20\%$ & 0.074 & 0.139 & 0.148 & 0.157 & 0.223 & 0.322 & 0.387 & 0.441 & 0.367 & 8.0 \\
$25\%$ & 0.074 & 0.139 & 0.139 & 0.167 & 0.231 & 0.333 & 0.366 & 0.446 & 0.373 & 7.4 \\
\textbf{30\%} & 0.134 & 0.171 & 0.227 & 0.227 & 0.249 & 0.348 & 0.419 & 0.486 & \textbf{0.418} & 7.6 \\
$35\%$ & 0.102 & 0.148 & 0.176 & 0.185 & 0.208 & 0.330 & 0.390 & 0.453 & 0.383 & 7.5 \\
\quad\emph{test} ($30\%$) & 0.102 & 0.153 & 0.173 & 0.194 & 0.200 & 0.305 & 0.347 & 0.414 & 0.357 & 7.6 \\
\bottomrule
\end{tabular}
\caption{Full training-ratio sweep for the two OOG strategies (\textbf{(1)}~the ``Other'' node and \textbf{(2)}~KG augmentation, defined in \S\ref{sec:ood}). Rows sweep the HV training OOG ratio with the HG fixed ($30\%$ and $35\%$, respectively); the validation-selected ratio ($25\%$/$30\%$, bold) is evaluated once on test (\emph{test} rows).}
\label{tab:ood_sweep_full}
\end{table*}

\section{Synthetic Dialogue Data Generation}
\label{app:synthetic_dialog_gen}
We introduce a latent \emph{knowledge graph grounding ratio} $\gamma \in [0,1]$, used only during synthetic dialogue generation and defined as the proportion of symptom-related information in a patient profile that maps to \texttt{symptom}, \texttt{cause}, or \texttt{risk\_factor} nodes in the diagnostic knowledge graph.

The thresholds on $\gamma$ set how strongly the model is instructed to rely on the oracle subgraph versus its internal knowledge: high-grounding ($\gamma \geq 0.7$) reasons primarily from the subgraph; moderate ($0.3 < \gamma < 0.7$) blends graph evidence with general clinical reasoning; and low-grounding ($\gamma \leq 0.3$) relies mainly on parametric knowledge.

This spectrum teaches the model when graph grounding is informative and when to fall back on internal knowledge. The complete system prompt is shown in Appendix~\ref{app:dialog_gen_prompts}.

\begin{table*}[t]
    \centering
    \scriptsize
    \resizebox{\textwidth}{!}{%
        \begin{tabular}{llcccccccc}
            \toprule
            \multirow{2}{*}{\textbf{Hypothesis Generator}} & \multirow{2}{*}{\textbf{Setting}} & \multicolumn{4}{c}{\textbf{Diagnostic Performance (Recall@$k$) $\uparrow$}} & \multirow{2}{*}{\textbf{Turns $\downarrow$}} & \multicolumn{2}{c}{\textbf{Last Turn Recall $\uparrow$}} \\
            \cmidrule(lr){3-6} \cmidrule(lr){8-9}
             & & \textbf{1} & \textbf{2} & \textbf{3} & \textbf{4} & & \textbf{HG} & \textbf{Subgraph} \\
            \midrule
            
            \textbf{Qwen2.5-7B-Instruct (Gen., FT)} & -- & 0.163 & 0.289 & 0.368 & 0.397 & 8.0 & 0.189 & 0.277 \\
            \midrule

            \multirow{6}{*}{\shortstack[l]{\textbf{SapBERT (Retriever)}\\\textbf{(Rerank O)}}} 
              & $n{=}1$ (20 dis) & 0.194 & 0.316 & 0.345 & 0.385 & 7.6 & 0.028 & 0.354 \\
              & $n{=}2$ (20 dis) & 0.184 & 0.311 & 0.366 & 0.394 & 7.8 & 0.080 & 0.385 \\
              & $n{=}3$ (20 dis) & 0.191 & 0.304 & 0.347 & 0.399 & \underline{7.3} & 0.083 & 0.354 \\
              & $n{=}4$ (20 dis) & \underline{0.222} & \underline{0.319} & \underline{0.370} & 0.396 & 8.0 & 0.132 & 0.368 \\
              & $n{=}5$ (20 dis) & 0.184 & 0.278 & 0.323 & 0.349 & 7.8 & 0.132 & 0.330 \\
            \midrule

            \multirow{6}{*}{\shortstack[l]{\textbf{SapBERT (Retriever)}\\\textbf{(Rerank X)}}} 
              & $n{=}1$ (20 dis) & 0.156 & 0.267 & 0.285 & 0.319 & 8.2 & 0.035 & 0.337 \\
              & $n{=}2$ (20 dis) & 0.163 & 0.280 & 0.318 & 0.349 & \underline{7.3} & 0.063 & 0.351 \\
              & $n{=}3$ (20 dis) & 0.191 & 0.286 & 0.335 & 0.361 & 7.8 & 0.111 & 0.354 \\
              & $n{=}4$ (20 dis) & 0.194 & 0.304 & 0.363 & \underline{0.413} & 7.7 & 0.146 & \underline{0.396} \\
              & $n{=}5$ (20 dis) & 0.184 & 0.288 & 0.330 & 0.359 & 7.7 & \underline{0.153} & 0.344 \\
            \midrule

            \multirow{1}{*}{\textbf{Qwen2.5-7B-Instruct + Head (Ours)}} 
              & $n{=}2, \tau{=}0.005$ & \textbf{0.250} & \textbf{0.361} & \textbf{0.394} & \textbf{0.418} & \textbf{6.9} & \textbf{0.258} & \textbf{0.559} \\
            \midrule

        \end{tabular}%
    }
    \caption{Detailed performance comparison of Hypothesis Generation methods. Best scores are highlighted in \textbf{bold}, and second-best scores are \underline{underlined}.}
    \label{tab:result_HG_approach_comparison}
\end{table*}

\section{Comparison of Hypothesis Generator Methods}
\label{app:HG_comparison}

\subsection{Generative Hypothesis Generator}
\label{app:HG_gen}
The generative approach fine-tunes a language model to produce candidate disease names directly from the dialogue history, without a classification head. For a fair comparison, it uses the same synthetic dialogue dataset as our classification-based HG.

Subgraph extraction follows the same multi-hop expansion as our main framework, using the top-$n$ generated diseases as anchors. Since the generative setting yields no calibrated probabilities for thresholding, we instead apply an attribute-overlap filter: a competing disease is retained only if the number of shared attribute nodes (\texttt{symptom}, \texttt{cause}, \texttt{risk\_factor}) with at least one anchor disease is at least 30\% of the anchor disease’s attribute set. This keeps diseases sufficiently related to the anchors while bounding subgraph size.

\subsection{Retrieval-Based Hypothesis Generator}
\label{app:HG_retrieval}
As an embedding-based alternative to the classification head, at each dialogue turn $t$ we use GPT-4o-mini \citep{openai2024gpt4omini} to extract the positive ($S_{pos}$) and negative ($S_{neg}$) symptoms mentioned by the patient, and linearize $S_{pos}$ into a query string. Each disease $d$ in the knowledge graph is represented by a canonical template listing its 1-hop attributes (\texttt{symptom}, \texttt{cause}, \texttt{risk\_factor}). We score query-disease embedding similarity with the biomedical encoder SapBERT \citep{liu-etal-2021-self}, which performed best in our preliminary experiments.

We evaluated two variations of this approach:
\begin{itemize}
        \item \textbf{Rerank X (No Rerank):} The system ranks diseases solely based on the cosine similarity between the positive symptom query and the disease document. The top-$n$ diseases are selected as anchor nodes.
    \item \textbf{Rerank O (With Rerank):} To penalize diseases associated with symptoms the patient has explicitly denied, we introduce a negative constraint. We linearize $S_{neg}$ and compute its similarity to each disease document $D_d$. The final score is calculated as:
    \begin{equation}
    \begin{split}
        Score(d) = \, & \text{Sim}(S_{pos}, D_d) \\
                      & - 0.3 \times \text{Sim}(S_{neg}, D_d)
    \end{split}
    \end{equation}
    Diseases are ranked by this adjusted score, and the top-$n$ are selected as anchor nodes.
\end{itemize}
As the extracted subgraph in our proposed setting ($n{=}2, \tau{=}0.005$) contains approximately 20 disease nodes on average, we standardized all retriever baselines by filtering the extracted subgraphs to retain the top-20 disease nodes based on embedding similarity for fair comparison.

\subsection{Comparison Result}
\label{app:HG_comparison_result}
As shown in Table~\ref{tab:result_HG_approach_comparison}, retrieval-based approaches underperform the classification head. Although \textit{Rerank O} improves Recall@1 over \textit{Rerank X} (0.222 vs. 0.194 at top 4), it still trails the classification head (0.250). Retrieval methods keep reasonable \textit{Subgraph Recall} yet low last-turn \textit{HG Recall}, indicating that embedding similarity alone struggles to pinpoint the exact disease among clinically similar conditions.

\section{Threshold Selection for CoD Baseline}
\label{app:cod_baseline}
Since CoD is originally designed to retrieve three diseases, we set its retrieval parameter to 20 and swept the confidence threshold ($\tau$) over 0.5, 0.55, and 0.6.

Table~\ref{tab:cod_threshold} summarizes the results. Across $\tau$ from 0.5 to 0.6, the average number of turns stays between 2.0 and 2.4, so the model terminates very early regardless of threshold. The $\tau=0.5$ setting achieves the highest Recall@1 (0.106), yet overall recall remains well below our SFT model.

We therefore report $\tau{=}0.5$ as the representative CoD setting in the main experiments, as it gives the best Recall@1.

\begin{table}[t]
    \centering
    \resizebox{\columnwidth}{!}{%
        \begin{tabular}{ccccccc}
            \toprule
            \multicolumn{2}{c}{\textbf{CoD Setting}} & \multicolumn{4}{c}{\textbf{Recall@$k \uparrow$}} & \\
            \cmidrule(lr){1-2} \cmidrule(lr){3-6}
            \textbf{Threshold ($\tau$)} & \textbf{Top $n$} & \textbf{1} & \textbf{2} & \textbf{3} & \textbf{4} & \textbf{Turns $\downarrow$} \\
            \midrule
            0.50 & 20 & \textbf{0.106} & \textbf{0.149} & \textbf{0.196} & 0.207 & \textbf{2.0} \\
            0.55 & 20 & 0.073 & 0.131 & 0.176 & \textbf{0.211} & 2.4 \\
            0.60 & 20 & 0.073 & 0.131 & 0.176 & \textbf{0.211} & 2.2 \\
            \bottomrule
        \end{tabular}%
    }
    \caption{Performance of the CoD baseline under different confidence thresholds ($\tau$) with fixed $n{=}20$. Best scores are in \textbf{bold}.}
    \label{tab:cod_threshold}
\end{table}

\section{Prompts for Synthetic Dialogue Generation}
\label{app:dialog_gen_prompts}

\begin{tcolorbox}[
  enhanced,
  breakable,
  colback=white,
  colframe=black,
  boxrule=0.8pt,
  arc=2mm,
  left=4mm,
  right=4mm,
  top=3mm,
  bottom=3mm,
  fontupper=\scriptsize, 
  coltitle=white,
  colbacktitle=black!80,
  fonttitle=\bfseries,
  attach boxed title to top left={xshift=0mm,yshift=-2mm},
  boxed title style={
    sharp corners,
    boxrule=0pt,
    colframe=black!80,
    colback=black!80,
    interior style={fill=black!80}
  }
]

You are a medical diagnosis assistant speaking only in ENGLISH.
You will simulate an ideal doctor who gradually narrows down the possible causes or makes a final diagnosis of a patient’s illness.\\

\textbf{Secret Background}\\
You (the AI) know the gold disease, but the simulated doctor does NOT.
Do not explicitly focus only on the gold disease.
Instead, reason like a real clinician who considers multiple possible diseases and gradually narrows them down through questioning.
Use the gold disease only to keep the reasoning medically consistent and to guide late-phase decisions.
Do NOT use it to jump prematurely to the diagnosis.\\

\textbf{Knowledge Graph Relevance (internal control)}\\
A hidden numerical score (ranging from 0 to 1) indicates how well the patient’s current symptoms are covered by the Knowledge Graph (KG).
You must implicitly balance the use of Knowledge Graph information and internal medical knowledge according to this score.
When the score is high ($\ge 0.7$), reasoning should be mostly grounded in the KG, while internal knowledge can fill small contextual gaps.
When the score is moderate ($0.3 < \text{score} < 0.7$), use both KG and internal knowledge in combination.
When the score is low ($\le 0.3$), rely mainly on internal medical knowledge, using the KG only if it still provides relevant hints.
Do not mention or reference this score explicitly in your <think> block or output.\\

\textbf{Action Generation Rules}\\
When deciding your next action, follow these structured steps.
You may either ask another question (Exploration) or make a final diagnosis (Exploitation) depending on the situation.
Do not delay diagnosis unnecessarily.
If your differential diagnosis has narrowed down to four possible diseases, make a working diagnosis immediately instead of continuing with more questions.
Even if additional tests or examinations might help in real life, you must still provide a reasonable diagnosis based on the available information.
Continue questioning only when the evidence is vague or conflicting and further information could meaningfully alter your reasoning.
If the dialogue length is \texttt{\{max\_turn\}} or more, you must make a final diagnosis without mentioning turn count or limits.\\

\textbf{1. Asking a Question (Exploration)}\\
\textit{Step 1. Examine the Knowledge Graph}\\
Begin every reasoning step by reviewing the KG.
Identify suspected diseases and the symptoms linked to them in the KG.
Consider which unexplored KG symptoms might refine the differential diagnosis.

\textit{Step 2. Check KG Coverage and Relevance}\\
Review what symptoms the patient has confirmed, denied, or is uncertain about.
Decide whether to ask about an unexplored KG symptom or ask a more informative question based on internal medical knowledge.

\textit{Step 3. Decide the Knowledge Source}\\
In your <think> block, you must list at least two suspected diseases, state whether you will use the KG or internal knowledge, and briefly justify why.

\textit{Step 4. Phrase the Question}\\
Choose exactly one unasked symptom or risk factor.
Avoid previously asked or semantically duplicate questions.
Keep the question short, simple, and patient-friendly.
Use one clear sentence.

\textit{Step 5. Output Format}\\
<think> ... </think> <question> your\_question </question>\\

\textbf{2. Making a Final Diagnosis (Exploitation)}\\
\textit{Step 1. Decide Whether to Diagnose}\\
Make a final diagnosis only when the differential diagnosis has narrowed to four possible diseases or when the dialogue length reaches \texttt{\{max\_turn\}} or more.

\textit{Step 2. Select the Disease(s)}\\
Select exactly four diseases.
\texttt{\{gold\_disease\}} must come first in the list.
All selected diseases must exist in the provided knowledge graph.

\textit{Step 3. Output Format}\\
<think> ... </think> <diagnosis> disease1, disease2, ... </diagnosis>\\

\textbf{Inputs}\\
Gold Disease (internal use only): \texttt{\{gold\_disease\}}\\
Knowledge Graph Relevance (internal use only): \texttt{\{relevance\_score\}}\\
Knowledge Graph: \texttt{\{subgraph\}}\\
Current Dialogue History: \texttt{\{dialogue\}}\\
Current Dialogue Length: \texttt{\{dialogue\_length\}}\\
Your Response:
\end{tcolorbox}

\section{Prompts for Patient Simulator}
\label{app:patient_prompts}

\subsection{Prompts used to induce low-specificity symptom descriptions}
\label{app:low_specificity_prompt}

\begin{tcolorbox}[
  enhanced,
  breakable,
  colback=white,
  colframe=black,
  boxrule=0.8pt,
  arc=2mm,
  left=4mm,
  right=4mm,
  top=3mm,
  bottom=3mm,
  fontupper=\scriptsize, 
  coltitle=white,
  colbacktitle=black!80,
  fonttitle=\bfseries,
  attach boxed title to top left={xshift=0mm,yshift=-2mm},
  boxed title style={
    sharp corners,
    boxrule=0pt,
    colframe=black!80,
    colback=black!80,
    interior style={fill=black!80}
  }
]

You play the role of a patient with 'Low Specificity'. You are unable to provide clear or precise details about your current symptom. You must adhere to the following guidelines for symptom description:

1. Location: Never pinpoint a specific spot. Use general terms like 'my belly', 'chest', or 'around here' instead of 'lower right quadrant' or 'under the rib'.

2. Character: Use vague, subjective adjectives like 'uncomfortable', 'weird', 'bad', or 'annoying'. STRICTLY AVOID medical or specific sensory terms like 'stabbing', 'burning', 'squeezing', or 'radiating'.

3. Duration and Onset: Do not use numbers or exact times. Use relative/fuzzy timeframes like 'a while ago', 'recently', 'not long', or 'since yesterday'.

4. Factors: Be unaware of what makes your symptom better or worse. Say 'I'm not sure', 'It just hurts all the time', or 'Nothing really changes it'.

\end{tcolorbox}

\subsection{Persona-conditional prompt components combined with the low-specificity prompt}
\label{app:compatibility_prompts}

\begin{tcolorbox}[
  enhanced,
  breakable,
  colback=white,
  colframe=black,
  boxrule=0.8pt,
  arc=2mm,
  left=4mm,
  right=4mm,
  top=3mm,
  bottom=3mm,
  fontupper=\scriptsize, 
  coltitle=white,
  colbacktitle=black!80,
  fonttitle=\bfseries,
  attach boxed title to top left={xshift=0mm,yshift=-2mm},
  boxed title style={
    sharp corners,
    boxrule=0pt,
    colframe=black!80,
    colback=black!80,
    interior style={fill=black!80}
  }
]
\label{box:specificity_compat}

\textbf{(Applied when persona includes High Recall)}\\
Your 'High Recall' applies only to medical history (past medical history, dates of past surgeries, etc.).
However, you must remain vague about your current symptoms.
For example, ``I took Amlodipine 5mg in the past, but my chest just feels weird today.''

\par\medskip\hrule\medskip

\textbf{(Applied when persona has High Language Proficiency)}\\
Even though you are fluent, do not describe symptoms in precise physical terms.
Instead, replace symptom descriptions with a self-diagnosed disease that you suspect or have researched.
If asked ``How does it feel?'', give a self-diagnosis rather than a description.
For example, instead of saying ``It is a squeezing chest pain'', say ``My chest hurts, I think it's angina.''

\par\medskip\hrule\medskip

\textbf{(Applied when persona is Verbose)}\\
Mask missing symptom details with long but largely irrelevant text.
Describe irrelevant contents in detail, but maintain a vague attitude regarding the actual symptom description.

\end{tcolorbox}

\section{Prompts for Hypothesis Generation}
\label{app:classifer_prompts}

\begin{tcolorbox}[
  enhanced,
  breakable,
  colback=white,
  colframe=black,
  boxrule=0.8pt,
  arc=2mm,
  left=4mm,
  right=4mm,
  top=3mm,
  bottom=3mm,
  fontupper=\scriptsize,
  coltitle=white,
  colbacktitle=black!80,
  fonttitle=\bfseries,
  attach boxed title to top left={xshift=0mm,yshift=-2mm},
  boxed title style={
    sharp corners,
    boxrule=0pt,
    colframe=black!80,
    colback=black!80,
    interior style={fill=black!80}
  }
]

Conversation:\texttt{\{dialogue\_history\}}

Based on the conversation, what diseases does the patient have?

\end{tcolorbox}

\section{Prompts for Hypothesis Verification}
\label{app:doctor_prompts}

\begin{tcolorbox}[
  enhanced,
  breakable,
  colback=white,
  colframe=black,
  boxrule=0.8pt,
  arc=2mm,
  left=4mm,
  right=4mm,
  top=3mm,
  bottom=3mm,
  fontupper=\scriptsize,
  coltitle=white,
  colbacktitle=black!80,
  fonttitle=\bfseries,
  attach boxed title to top left={xshift=0mm,yshift=-2mm},
  boxed title style={
    sharp corners,
    boxrule=0pt,
    colframe=black!80,
    colback=black!80,
    interior style={fill=black!80}
  }
]

You are a medical diagnosis assistant speaking only in ENGLISH.

Your goal is to decide whether to ask another question or make a final diagnosis,
based on the current knowledge graph and dialogue history.
Combine the medical knowledge you already have with the provided graph and dialogue.

\medskip
\textbf{Decision Rules}

\textbf{1. Ask another question (exploration)}

-  Ask a question when there are still important unasked symptoms or risk factors in the knowledge graph and additional information could meaningfully change the diagnosis.

-  Select exactly one unasked symptom or risk factor that best distinguishes between diseases.

-  Do not ask about diseases directly; ask only about symptoms or findings.

-  Avoid asking semantically duplicate questions already present in the dialogue history.

-  Use patient-friendly language without medical jargon.

-  Internal knowledge may be used to ask exactly one high-yield symptom not present in the graph when clinically necessary.

- \textbf{Output format:} <question> your\_question </question>

\medskip
\textbf{2. Make a final diagnosis (exploitation)}

-  Make a diagnosis when the patient’s symptoms clearly match a disease in the graph or when further questions are unlikely to change the conclusion.

-  Select up to four diseases from the knowledge graph.

- \textbf{Output format:} <diagnosis> disease1, disease2, ... </diagnosis>

\medskip
\textbf{3. Action constraint}\\
Return exactly one action: either <question> or <diagnosis>.

\medskip
\textbf{4. Reasoning and output format}\\
First write reasoning inside <think> ... </think>, using both the knowledge graph and internal medical knowledge.
If internal knowledge is used, briefly note this inside <think>.
Do not output explanations outside the <think> block.

\medskip
\textbf{Inputs}

- Knowledge Graph: \texttt{\{subgraph\}}

- Current Dialogue History: \texttt{\{dialogue\}}

\textbf{Your Response:}
\end{tcolorbox}

\section{Examples of Hypothesis Verifier Outputs}
\label{app:doctor_response_examples}

\begin{tcolorbox}[
  enhanced,
  breakable,
  colback=white,
  colframe=black,
  boxrule=0.8pt,
  arc=2mm,
  left=4mm,
  right=4mm,
  top=3mm,
  bottom=3mm,
  fontupper=\scriptsize, 
  coltitle=white,
  colbacktitle=black!80,
  fonttitle=\bfseries,
  attach boxed title to top left={xshift=0mm,yshift=-2mm},
  boxed title style={
    sharp corners,
    boxrule=0pt,
    colframe=black!80,
    colback=black!80,
    interior style={fill=black!80}
  }
]

\textbf{Example 1: Clarifying Question Generation}

\texttt{<think>} 
Based on the patient's symptoms of lower right abdominal pain, decreased appetite, and fatigue, several potential conditions could be considered. The differential diagnosis includes Crohn's disease, acute appendicitis, infectious mononucleosis, and possibly a breast abscess if there were any related symptoms. Given the patient's lack of nausea or vomiting and the presence of diarrhea, Crohn's disease seems particularly relevant due to its association with gastrointestinal symptoms. Since the Knowledge Graph provides insights into these conditions, I will focus on exploring symptoms that could help narrow down the diagnosis further.
\texttt{</think>}
\texttt{<question>} 
Have you noticed any changes in your bowel movements, like diarrhea or constipation?
\texttt{</question>}

\medskip
\hrule
\medskip

\textbf{Example 2: Final Diagnosis Generation}

\texttt{<think>} 
Based on the patient's symptoms of chest tightness, shortness of breath, leg swelling, and abdominal discomfort, several potential conditions could be considered. These include congestive heart failure, atrial fibrillation, chronic obstructive pulmonary disease (COPD), and possibly pericarditis. Given the patient's presentation, particularly the leg swelling and abdominal fullness, congestive heart failure is a strong contender. The Knowledge Graph supports this with symptoms like dyspnea, edema, and abdominal fullness being linked to heart failure. Since the dialogue has reached a sufficient length, I will proceed to make a final diagnosis.
\texttt{</think>}

\texttt{<diagnosis>}  
congestive heart failure, atrial fibrillation, chronic obstructive pulmonary disease (COPD), pericarditis
\texttt{</diagnosis>}
\end{tcolorbox}

\clearpage

\section{Error Analysis}
\label{app:Error_Analysis}

Our qualitative analysis of misdiagnosed cases found that discrepancies between textbook definitions and real-world patient narratives, together with noise in EHR-derived labels, drove most errors; these fall into four primary types.
We further quantified their frequency by manually categorizing all 162 misdiagnosed cases (Recall@4 ${=}$ 0); Table~\ref{tab:error_freq} reports the counts. The four primary types account for the majority; the remainder corresponds to adjacent-differential confusion and a few minor categories (multi-disease partial recognition, organ-system routing failure, cancer/metastasis recognition gap, and sepsis-source confusion). Notably, the label-granularity errors (e.g., \emph{heart failure} vs.\ \emph{congestive heart failure}) are frequently top-1 yet scored as misses under exact string matching, which partly explains why Recall@4 remains below 50\%.

\paragraph{1. Subjectivity and Imprecision in Patient Narratives}
Textbooks describe diseases with prototypical symptom patterns (e.g., ``right lower quadrant pain'' for acute appendicitis), but real-world patients often struggle to localize pain or describe symptoms precisely.
\begin{itemize}
    \item \textbf{Case 1:} The patient had \textit{Acute Appendicitis} but described the pain location as ``in the upper part'' of the stomach. Consequently, the model predicted upper-abdominal pathologies such as \textit{Acute Cholecystitis} or \textit{Gallstones}.
    \item \textbf{Case 2:} The patient reported abdominal pain after eating spaghetti (fatty food), leading the model to predict \textit{Diverticulitis} or \textit{IBS}. The ground truth was \textit{Chronic Pancreatitis}. While the trigger (eating) was consistent, the vague description of the pain led to misdiagnosis.
\end{itemize}

\paragraph{2. Misinterpretation of Atypical Symptoms in Specific Populations}
Systemic symptoms such as falls, confusion, or weakness are common atypical presentations of infections (e.g., UTI, Pneumonia) in the elderly, yet the model frequently misinterpreted them as neurological events.
\begin{itemize}
    \item \textbf{Case 1:} The patient reported, ``I fell down. Just confused and tired,'' with no specific urinary complaints. The model interpreted the fall and confusion as signs of a \textit{Transient Ischemic Attack (TIA)} or \textit{Parkinson's Disease}. The ground truth was a \textit{Urinary Tract Infection (UTI)}, a condition known for causing delirium and falls in geriatric patients without typical dysuria.
    \item \textbf{Case 2:} Similarly, a patient reporting a fall, confusion, and speech difficulty was diagnosed with \textit{Pneumonia}, but the model predicted \textit{Cerebral Infarction} or \textit{TIA}. This highlights the challenge of diagnosing systemic infections based solely on dialogue when localizing symptoms are absent.
\end{itemize}

\begin{table}[t]
\centering
\small
\resizebox{\columnwidth}{!}{%
\begin{tabular}{lcc}
\toprule
Error type & Count & \% of misdiag. \\
\midrule
Subjective / vague patient narratives & 40 & 24.7\% \\
Acute vs.\ chronic confusion          & 33 & 20.4\% \\
Label granularity (sibling diseases)  & 28 & 17.3\% \\
Elderly atypical presentation         & 21 & 13.0\% \\
\bottomrule
\end{tabular}}
\caption{Frequency of the four error types among the 162 misdiagnosed cases. The remainder corresponds to adjacent-differential confusion and four minor categories.}
\label{tab:error_freq}
\end{table}

\paragraph{3. Confusion Between Acute and Chronic Conditions}
Respiratory symptoms such as dyspnea, cough, and chest tightness are shared by acute infections and chronic lung diseases. Absent distinct acute markers (e.g., sudden high fever or purulent sputum) in the dialogue, the model often defaulted to chronic diagnoses.
\begin{itemize}
    \item \textbf{Case 1:} The patient reported ``trouble breathing'' and ``coughing a lot.'' Although these are classic signs of \textit{Pneumonia} (Ground Truth), the model interpreted them as indicators of chronic airway diseases like \textit{Asthma} or \textit{Chronic Bronchitis}.
    \item \textbf{Case 2:} The patient described chest pressure and a cough persisting for ``two weeks.'' This duration likely introduced ambiguity, leading the model to prioritize chronic conditions such as \textit{COPD} over the correct acute diagnosis of \textit{Pneumonia}.
\end{itemize}

\paragraph{4. Label Granularity}
In several cases the model's reasoning was pathophysiologically sound, but the prediction was counted as a failure due to a label-granularity (hierarchy) mismatch.
\begin{itemize}
    \item \textbf{Case 1:} The patient reported heart racing and weakness. The model correctly identified the broader category \textit{Arrhythmia}, but the ground truth was the specific subtype \textit{Atrial Flutter}. Since \textit{Arrhythmia} and \textit{Atrial Flutter} exist as distinct nodes in our KG, this was penalized despite the reasoning being directionally correct.
\end{itemize}

\section{Detailed Analysis of Patient Simulator Evaluation} \label{sec:appendix_human_eval}

\subsection{Evaluation Protocol}
\label{app:patient_simulator_evaluation_details}
For each case, physicians saw the full dialogue history and the two alternative patient responses, then answered the pairwise preference question, \textit{``Which response is more representative of patients encountered in actual clinical practice?''}, with a ``Tie'' option when the two were indistinguishable. 
This setup provides a controlled response-level comparison under the same dialogue context. It is also conservative for our simulator, since the preceding dialogue history was generated by the baseline simulator rather than by our specificity-augmented simulator.

Despite this disadvantage, we aimed to test whether adding low-specificity behavior produces more realistic patient responses.

\subsection{Qualitative Feedback from Physicians}
\label{app:patient_simulator_comment}
Physician feedback clarified the nuances of these preferences:
\begin{itemize}
    \item \textbf{Physician 1:} Emphasized that real patients usually have a basic interest in their symptoms and can provide some level of localization. They preferred our system because it avoided the extremes of being "too indifferent/lazy" or "too professional/medically knowledgeable," both of which were perceived as unrealistic.
    \item \textbf{Physician 2:} Noted that their expertise influenced the evaluation. As a specialist in gastrointestinal (GI) diseases, they found our simulator's vague responses highly realistic for GI cases. Conversely, for respiratory or circulatory cases not encountered in many years, they found it harder to distinguish typical patient types, leading to a higher frequency of "comparable (Tie)" ratings.
    \item \textbf{Physician 3:} Focused on the overall alignment with general patient behavioral patterns seen in daily practice.
\end{itemize}

Post-evaluation interviews identified the primary criteria for judging simulator naturalness:
\begin{enumerate}
    \item \textbf{Realistic Vagueness:} Two out of three physicians identified "Vagueness" as the most critical factor. Even without explicit instructions, they favored the simulator that provided vague but plausible information, as it mimics how real patients often struggle to describe symptoms precisely.
    \item \textbf{Engagement and Naturalness:} Responses that appeared overly indifferent or lacked effort were penalized. The physicians felt that while patients may be vague, they are rarely completely uncooperative, and our system better captured this balance compared to the baseline.
\end{enumerate}

\subsection{Inter-Rater Agreement}
\label{app:patient_simulator_agreement}
Table~\ref{tab:irr} summarizes inter-rater agreement. On the full 41 scenarios, agreement is modest because one physician (Physician~1, a gastroenterology specialist) marked 22 of 41 scenarios as ``comparable''---predominantly non-GI presentations outside their primary expertise---rather than committing to a preference. Because the remaining preferences are heavily skewed toward our simulator, Fleiss' $\kappa$ is distorted by the prevalence paradox (Feinstein \& Cicchetti, 1990) and is near zero ($-0.06$) despite substantial raw agreement; we therefore also report Gwet's AC1 (Gwet, 2008), which is robust to prevalence. Restricting to the 19 scenarios where Physician~1 gave a definitive judgment, pairwise agreement and AC1 both rise sharply, and all three physicians prefer our simulator far more often, indicating that disagreement reflects specialty-limited confidence rather than genuine conflict over patient realism.

\begin{table}[t]
\centering
\small
\setlength{\tabcolsep}{4pt}
\resizebox{\columnwidth}{!}{%
\begin{tabular}{lcc}
\toprule
Metric & All 41 & Committed 19 \\
\midrule
Pref.\ ours: Physician~1 (GI)    & 43.9\% & 94.7\% \\
Pref.\ ours: Physician~2         & 58.5\% & 73.7\% \\
Pref.\ ours: Physician~3 (Surg.) & 51.2\% & 57.9\% \\
\midrule
Agreement P1--P2 & 34.1\% & 68.4\% \\
Agreement P2--P3 & 41.5\% & 42.1\% \\
Agreement P1--P3 & 29.3\% & 63.2\% \\
\midrule
Gwet's AC1 & 0.06 & 0.48 \\
Fleiss' $\kappa$ & $-0.06$ & (paradox) \\
\bottomrule
\end{tabular}}
\caption{Inter-rater agreement among the three physicians, over all 41 scenarios vs.\ the 19 scenarios where Physician~1 (GI specialist) gave a definitive (non-``comparable'') judgment. Fleiss' $\kappa$ is uninformative here due to the prevalence paradox; Gwet's AC1 is prevalence-robust.}
\label{tab:irr}
\end{table}

\section{Detailed Analysis of CQ Quality Evaluation}
\label{app:cq_eval_details}
\subsection{Evaluation Protocol}
Evaluation used a web-based platform: physicians role-played patients, responding to the HV's questions using only the clinical facts in each profile, then retrospectively scored the full dialogue history.

\subsection{Evaluation Metrics}
\begin{itemize}
    \item \textbf{Essentiality:} Measures whether the questions were necessary to rule in or out candidate diseases, or if they were redundant/irrelevant.
    \item \textbf{Conversational Flow:} Evaluates the logical transition between questions, assessing if the system follows a natural diagnostic narrative (e.g., from chief complaint to associated symptoms).
    \item \textbf{Clinical Authenticity:} Assesses whether the content of CQs align with the professional standards of actual history-taking.
\end{itemize}

\subsection{Qualitative Feedback from Physicians}
Physicians offered several constructive comments:
\begin{itemize}
    \item \textbf{Physician 1:} Emphasized the importance of thorough inquiry into "red flags" (e.g., weight loss, specific bleeding patterns) to rule out critical conditions like malignancy. They suggested that the system could further improve by asking more detailed questions regarding pain characteristics, such as radiation and migration, particularly for complex abdominal cases.
    \item \textbf{Physician 2:} Noted that while the diagnostic approach was generally logical, the system could benefit from placing greater weight on specific patient-reported past medical history (e.g., prior cardiovascular events) and vital sign indicators when prioritizing the differential diagnosis list.
    \item \textbf{Physician 3:} Commended the system for its stable and accurate diagnostic workflow. They assessed the question sequences as natural and aligned with standard clinical reasoning protocols.
    \item \textbf{Physician 4:} Suggested improving conversation efficiency by combining related questions to reduce redundancy. Additionally, they noted that incorporating multi-modal capabilities (e.g., viewing skin lesion images) and enhancing responsiveness to direct patient cues would further increase clinical utility.
\end{itemize}

\subsection{Physician Free-Text Assessment}
\label{app:freetext_assessment}
Beyond the interactive Likert study above, two board-certified physicians (clinical pharmacology and hematology-oncology) each read ten full inference dialogues and commented in free text on clarifying-question quality and diagnostic sufficiency (their patient-realism observations are summarized in \S\ref{sec:patient_simulator_eval}). Both found the clarifying questions only partially appropriate, missing vital signs, red-flag and severity assessment, and focused history, with occasional repetition or premature closure (``settled on anorexia nervosa too quickly''). Both also judged a text-only history intrinsically insufficient for a definitive diagnosis when physical examination or imaging is required (e.g., ulcerative colitis needs colonoscopy), positioning the system as useful for forming an initial impression (``usable at an educational level'') rather than definitive diagnosis.

\end{document}